%% file: main.tex
\pdfoutput=1

\documentclass[11pt]{article}

\usepackage[]{acl}

\usepackage{times}
\usepackage{latexsym}

\usepackage[T1]{fontenc}

\usepackage[utf8]{inputenc}

\usepackage{tcolorbox}
\tcbuselibrary{skins, breakable, listings}
\usepackage{xcolor}
\usepackage{colortbl}
\usepackage{tabu}
\usepackage{booktabs}
\usepackage{multirow}

\usepackage{graphicx} 
\usepackage{float}
\usepackage{subfigure} 
\usepackage{amssymb}

\usepackage{hyperref}
\usepackage{tablefootnote}
\usepackage{xspace}

\usepackage{float}
\usepackage{amsmath}
\usepackage{bm}
\usepackage{algorithmicx}
\usepackage{algorithm}
\usepackage{algpseudocode}

\usepackage{arydshln}

\usepackage{amssymb}
\usepackage{pifont}
\newcommand{\cmark}{\ding{51}}%
\usepackage{enumitem}

\usepackage{fontawesome5}

\usepackage{microtype}

\newcommand{\ours}{\textsc{MISS-QA}\xspace}
\newcommand{\qasper}{\textsc{Qasper}\xspace}

\newcommand{\testmini}{\emph{testmini}\xspace}
\newcommand{\test}{\emph{test}\xspace}
\newcommand{\lmm}{multimodal foundation model\xspace}
\newcommand{\eg}{\hbox{\emph{e.g.,}}\xspace}

\newcommand{\ie}{\hbox{\emph{i.e.,}}\xspace}

\newcommand{\qnum}{1,500\xspace}
\newcommand{\pnum}{465\xspace}
\newcommand{\nmodel}{18\xspace}
\newcommand{\npeople}{16\xspace}

\newcommand{\papers}{scientific papers\xspace}
\newcommand{\paper}{scientific paper\xspace}
\newcommand{\fig}{schematic diagram\xspace}
\newcommand{\figs}{schematic diagrams\xspace}

\definecolor{Gray}{gray}{0.95}
\newcolumntype{a}{>{\columncolor{Gray}}c}

\usepackage{listings}

\definecolor{customblue}{HTML}{286dc0}

\definecolor{customred}{HTML}{d62728}

\definecolor{OliveGreen}{rgb}{0.33, 0.42, 0.18}

\newtcolorbox{blueBox}[1][]{
  colback=customblue!5!white,
  colframe=customblue,
  floatplacement=floating,
  title=\centering #1
}

\newtcolorbox{greyBox}[1][]{
  colback=customgrey!10!white,
  colframe=customgrey,
  floatplacement=floating,
  title=\centering #1
}

\newtcolorbox{wronganswer}[1][]{
    enhanced,
    breakable,
    colframe=customred,
    colback=customred!10!white,
    sharp corners,
    boxsep=0pt,
    left=5pt,
    right=5pt,
    top=6pt,
    bottom=6pt,
    boxrule=0pt,
    leftrule=4pt,
    #1
}

\newtcolorbox{correctanswer}[1][]{
    enhanced,
    breakable,
    colframe=OliveGreen,
    colback=OliveGreen!10!white,
    sharp corners,
    boxsep=0pt,
    left=5pt,
    right=5pt,
    top=6pt,
    bottom=6pt,
    boxrule=0pt,
    leftrule=4pt,
    #1
}

\newcommand{\huggingface}{\raisebox{-1.5pt}{\includegraphics[height=1.05em]{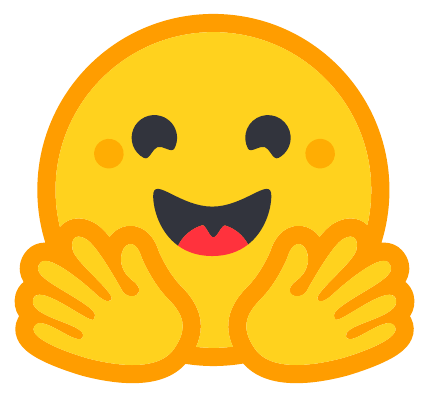}}\xspace}
\newcommand{\github}{\raisebox{-1.5pt}{\includegraphics[height=1.05em]{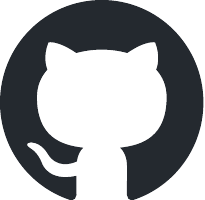}}\xspace}

\addtocontents{toc}{\protect\setcounter{tocdepth}{-1}}
\title{Can Multimodal Foundation Models Understand Schematic Diagrams?
\\An Empirical Study on Information-Seeking QA over Scientific Papers}

\author{Yilun Zhao\thanks{~~Equal Contributions. Correspondence: Yilun Zhao (\texttt{yilun.zhao@yale.edu})} \quad Chengye Wang$^*$ \quad Chuhan Li \quad Arman Cohan \vspace{5pt}\\
Yale NLP Lab}

\begin{document}
\maketitle
\begin{abstract}
\input{main/0-abstract}
\end{abstract}

\input{main/1-introduction}
\input{main/2-related_work}
\input{main/3-task}
\input{main/4-2-experiment}
\input{main/5-conclusion}
\input{main/limitations}


\bibliography{anthology,custom,llms}
\appendix

\addtocontents{toc}{\protect\setcounter{tocdepth}{3}}

\renewcommand{\contentsname}{\large Appendix Contents}
\hypersetup{linkcolor=black}
\tableofcontents
\newpage
\input{appendix/a1}

\end{document}

%% file: main/0-abstract.tex
This paper introduces MISS-QA, the first benchmark specifically designed to evaluate the ability of models to interpret schematic diagrams within scientific literature. 
MISS-QA comprises 1,500 expert-annotated examples over 465 scientific papers. 
In this benchmark, models are tasked with interpreting schematic diagrams that illustrate research overviews and answering corresponding information-seeking questions based on the broader context of the paper. 
We assess the performance of 18 frontier multimodal foundation models, including o4-mini, Gemini-2.5-Flash, and Qwen2.5-VL. We reveal a significant performance gap between these models and human experts on MISS-QA. Our analysis of model performance on unanswerable questions and our detailed error analysis further highlight the strengths and limitations of current models, offering key insights to enhance models in comprehending multimodal scientific literature.

\begin{small}
\begin{center}
\begin{tabular}{ll}
\huggingface~~\textbf{Data} &  \href{https://huggingface.co/datasets/yale-nlp/MISS-QA} {\path{yale-nlp/MISS-QA}}\\
\github~~\textbf{Code} &\href{https://github.com/yilunzhao/MISS-QA}{\path{yilunzhao/MISS-QA}}\\
\end{tabular}
\end{center} 
\end{small}

%% file: main/1-introduction.tex
\input{figure_tex/main_example}
\input{tables/dataset_comparison}

\section{Introduction}
When engaging with academic literature, readers frequently first consult \figs that illustrate the research framework
to quickly grasp the underlying research and identify specific areas of interest. 
This initial visual reference is crucial as it provides an overview of the study's structure and aims, prompting readers to delve deeper into the paper content for a more comprehensive understanding. 
Given the complexity and high volume of information in \papers~\cite{dasigi-etal-2021-dataset, pmlr-v202-lee23n, li-etal-2024-m3sciqa, ajith-etal-2024-litsearch, zhao2025sciarena, xu2025llmsidentifycriticallimitations, wang2025sciver}, there is a significant demand for AI systems that can assist scientists in seeking information from paper through \figs.

Foundation models have recently demonstrated remarkable capabilities in tackling various complex tasks~\cite{Touvron2023Llama2O, Jiang2023Mistral7, openai2024gpt4o, Zhao_2025_CVPR}.
However, their capabilities in interpreting \figs and synthesizing information from related paper context remain largely underexplored. 
Previous scientific document QA benchmarks primarily focus on text-only contexts~\cite{dasigi-etal-2021-dataset, pmlr-v202-lee23n}, emphasizing information-seeking questions derived from paper abstracts.
While recent scientific multimodal QA benchmarks have incorporated multimodal information from scientific papers~\cite{li-etal-2024-m3sciqa, wang2024charxiv, li2024multimodal, li2024mmsci}, their focus is largely limited to analyzing charts or tables that present experimental results.
In contrast, reasoning over schematic diagrams introduces unique challenges. 
It requires understanding not only the visual elements but also their relationships, contextual meaning, and integration with the textual narrative of the paper. 
Addressing these challenges is critical for advancing AI systems to effectively navigate the multimodal nature of scientific research.

To bridge this gap, we introduce \textbf{\ours}, a question answering benchmark focuses on \textbf{\underline{M}}ultimodal \textbf{\underline{I}}nformation-\textbf{\underline{S}}eeking over \textbf{\underline{S}}cientific papers.
The benchmark comprises \qnum expert-annotated QA examples over \pnum \papers, capturing a wide range of real-world information-seeking scenarios over scientific literature.
For each question, the schematic diagram—typically appearing as the first or second figure in the paper and providing an overview of the research—serves as the focal point. Foundation models are tasked with identifying the highlighted visual element within the diagram and answering the question using the content of the paper.
To mirror real-world information-seeking scenarios, where researchers often encounter questions that cannot be answered using the available context, 26.5\% of the questions in \ours are intentionally designed to be unanswerable~\cite{asai-choi-2021-challenges, sulem-etal-2022-yes}. Such questions tests the model's ability to discern the limits of the provided information, which is crucial for scientific research.

\autoref{fig:example} illustrates the key research questions investigated in this study.
Our main contributions are summarized as follows:

\begin{itemize} [leftmargin=*]
\itemsep0em
\item We present \ours, the first benchmark specifically designed to assess the ability of foundation models to comprehend schematic diagrams in scientific literature.

\item We perform a comprehensive evaluation of \nmodel frontier multimodal foundation models, uncovering a significant performance disparity between human experts and current models.

\item We analyze model performance on unanswerable questions and conduct an in-depth error analysis, highlighting key strengths, limitations, and opportunities to improve model performance.

\end{itemize}

%% file: figure_tex/main_example.tex
\begin{figure}[!t]
    \centering
    \includegraphics[width = \linewidth]{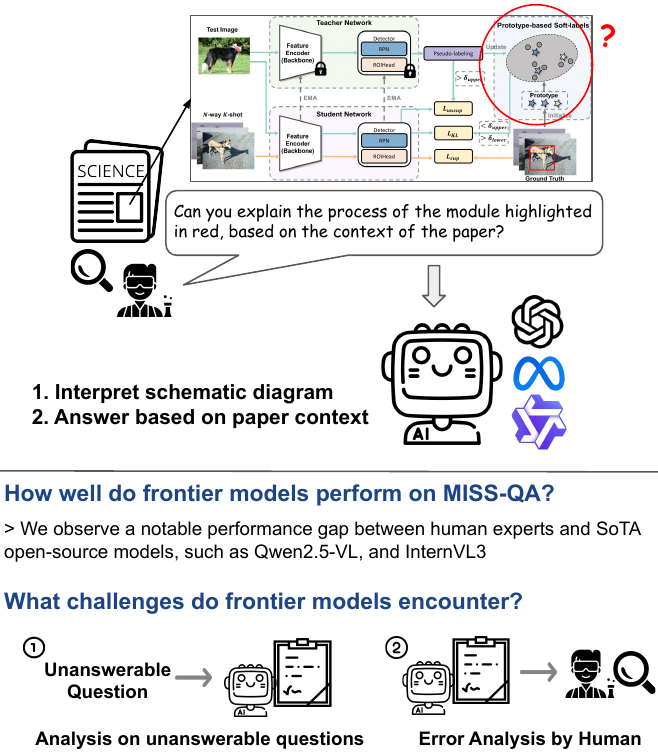}
    \caption{An overview of the \ours benchmark and research questions investigated in this study.
    }
    \label{fig:example}
\end{figure}

%% file: tables/dataset_comparison.tex
\begin{table*}[!t]
\centering
\renewcommand{\arraystretch}{1.2}
\small
\begin{tabular}{lll}
\hline
\toprule
\textbf{Dataset}     & \textbf{Input Context} & \textbf{Annotation / DatSingle Creation} \\ 
\midrule
\multicolumn{3}{c}{\textbf{\emph{Scientific Figure Understanding}}} \\
\noalign{\vskip 1ex}
ArXivQA~\cite{li-etal-2024-multimodal-arxiv} & Single scientific chart for experiment results & Rewritten from caption by GPT-4V \\
MMSci~\cite{li2024mmsci} & Single scientific chart for experiment results & Rewritten from caption by experts \\
CharXiv~\cite{wang2024charxiv} & Single scientific chart for experiment results & Annotated by human experts\\
\midrule
\multicolumn{3}{c}{\textbf{\emph{Scientific Paper QA}}} \\
\noalign{\vskip 1ex}
\qasper~\cite{dasigi-etal-2021-dataset} & Single text-only scientific paper & Annotated by human experts \\
QASA~\cite{qasa} & Single text-only scientific paper & Annotated by human experts \\
M3SciQA~\cite{li-etal-2024-m3sciqa} & Multiple scientific papers with charts & Synthesized by GPT-4V\\
\midrule
\textbf{\ours (ours)} & Single schematic diagram and corresponding paper & Annotated by human experts \\ 
\bottomrule
\hline
\end{tabular}

\caption{
Comparisons between \ours and existing scientific figure understanding and paper QA benchmarks.
}
\label{tab:dataset_comparison}
\end{table*}

%% file: main/2-related_work.tex
\input{figure_tex/pipeline}
\section{Related Work}
Early document QA benchmarks~\cite{rajpurkar2016squad, bajaj2018msmarco, yang2018hotpotqa} target short documents (\eg Wikipedia pages) and single-hop reasoning.
In contrast, question answering over scientific papers introduces unique challenges. These include interpreting domain-specific terminology, synthesizing information across multiple sections, and reasoning over multimodal content. 
However, as shown in \autoref{tab:dataset_comparison}, existing benchmarks in this area mainly investigate the \emph{text-only} scenarios. 
Specifically, \qasper~\cite{dasigi-etal-2021-dataset} focuses on tasks where questions are derived from the abstract of a \paper, with answers found within its main sections. Similarly, QASA~\cite{pmlr-v202-lee23n} requires more in-depth reasoning over the paper to answer the questions, yet it also focuses solely on text-based context. 
The most recent advancement, M3SciQA~\cite{li-etal-2024-m3sciqa}, expands scientific paper QA to include multimodal contexts, focusing specifically on interpreting charts that present experimental results. Parallel efforts in scientific figure understanding~\cite{li-etal-2024-multimodal-arxiv, wang2024charxiv, li2024mmsci} similarly emphasize reasoning over charts. However, these benchmarks overlook other critical visual modalities such as schematic diagrams, which play a central role in many scientific disciplines.

%% file: figure_tex/pipeline.tex
\begin{figure*}[!t]
\centering
\includegraphics[width=1\textwidth]{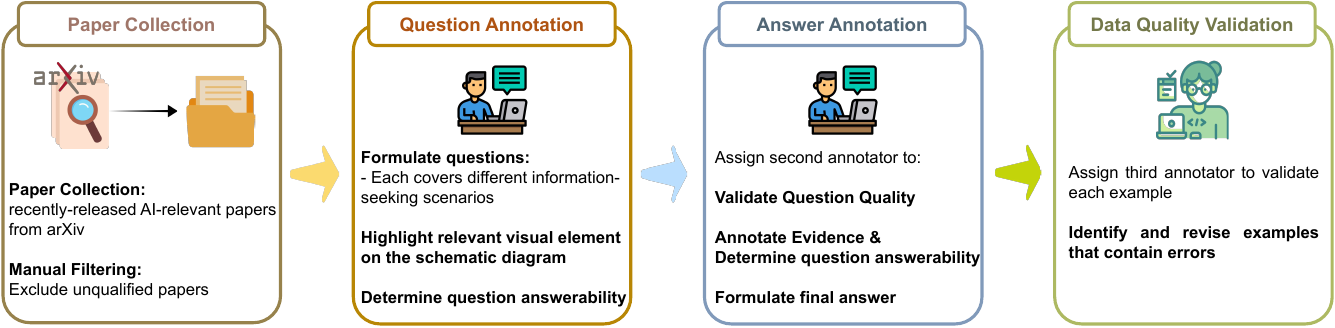}
\caption{An overview of the \ours benchmark construction pipeline.
}
\label{fig:pipeline}
\end{figure*}

%% file: main/3-task.tex
\section{The \ours Benchmark}\label{sec:data}
We develop the \ours benchmark to evaluate foundation models' abilities to interpret \figs and integrate them into the information-seeking process over scientific papers. We formally define the task of \ours in the context of foundation models (FM) as follows:
Given a \fig $diag$ within a \paper, an associated question $q$, and the main sections of paper $p$, the task is to understand and interpret $diag$, identify the question-relevant information within $p$, and generate the final free-form answer $a$:
\begin{equation}
\label{eq:formulation}
    \hat{a} = \arg\max_a P_{\mathbf{FM}}(a~|~diag, q, p)
\end{equation}

\noindent The following subsections outline the preliminary setup for data construction, followed by a comprehensive explanation of the benchmark construction and validation processes.
\autoref{fig:pipeline} presents an overview of the data construction pipeline.

\subsection{Preliminary Setup}\label{sec:data-setup}
We first discuss the preliminary setup.

\paragraph{Benchmark Construction Desiderata.}
To ensure the high quality of data, \ours adheres to the following three benchmark construction desiderata:
(1) \emph{Real-world research scenarios}: it mirrors real-world information-seeking scenarios in which researchers interact with scientific literature, emphasizing the use of \figs to swiftly comprehend and navigate intricate information within a scientific paper. 
(2) \emph{Schematic diagram interpretation requirement}: it demands the interpretation of \figs, ensuring that models do more than merely scan text for answers—they must genuinely interpret and integrate visual information for reasoning;
and (3) \emph{Diverse reasoning types}: it includes a range of reasoning types to reflect the complexity of information-seeking questions that researchers encounter when critically reading the paper context.

\paragraph{Expert Annotator Recruitment and Training.}
Given the high level of expertise required for \ours, annotation could not be entrusted to crowdsourced annotators. Instead, we recruit \npeople human researchers as expert annotators. 
Their anonymized profiles are detailed in Appendix~\ref{sec:annotator-info}. Notably, each annotator has authored a minimum of three AI-related publications, ensuring a high standard of domain knowledge.
\ours primarily targets \papers across diverse subfields within the \emph{AI domain} (\eg natural language processing, computer vision, and machine learning), aligning with the expertise of our annotators. 
Before commencing the annotation process, each annotator undergoes a comprehensive one-hour training session conducted by one of the authors. 
The screenshots of our developed annotation interface are presented in Appendix~\ref{app:interface}.

\input{figure_tex/figure3}
\paragraph{Information-Seeking Scenario and Subset Categorization.}
To better understand the common information-seeking scenarios researchers encounter when reading scientific papers, we conducted a preliminary study. In this study, annotators were given the freedom to read any scientific paper and annotate any type of question that seeks information contained within the context of the paper. This process allowed us to identify four key types of scenarios that frequently arise in practical settings.
Based on these findings, we created five subsets of \ours, each tailored to specific information-seeking focuses: design rationale, implementation details, literature background, experimental results, and others. We define each subset and provide corresponding example questions in \autoref{tab:data-type} and Appendix~\ref{app:example}, respectively.

\subsection{Source Paper Collection and Filtering}
We next outline the process of collecting and filtering source papers.

\paragraph{Source Paper Collection} 
We collect AI-related scientific papers from arXiv, focusing on subfields that align with the expertise of our annotators. Specifically, we select papers categorized under ``Artificial Intelligence'', ``Computation and Language'', ``Machine Learning'', ``Computer Vision and Pattern Recognition'' and ``Information Retrieval''. 
Our dataset includes papers first released between July 1 and November 30, 2024.
This time frame is after the cutoff date for most open-source pretraining corpora used to train foundation models, helping to mitigate issues related to data memorization in the models evaluated in this study~\cite{deng-etal-2024-unveiling}.
For each paper, we extract its context, including figures and tables, from the HTML\footnote{\url{https://info.arxiv.org/about/accessible_HTML.html}}
version provided by the arXiv platform. 
This process results in a total of \pnum \papers.

\paragraph{Source Paper Filtering} 
Five expert annotators (\ie one for each subfield) are then assigned to manually filter out unqualified papers. 
Specifically, we exclude (1) papers that are not focused on research (\eg surveys, position papers, and dissertations), and thus is difficult to design research-related questions; (2) papers whose first two figures are both not \figs or are unsuitable for composing challenging questions that requires figure interpretation; and (3) papers that expert annotators judged to be of low quality or that contained errors in paper parsing based on their assessment.
This results in \pnum papers for subsequent annotation.
For each paper collected, the annotators are tasked with labeling the first figure identified as a schematic diagram, which is then used for subsequent QA annotation.

\begin{figure}[!t]
    \centering
    \includegraphics[width = 0.98\linewidth]{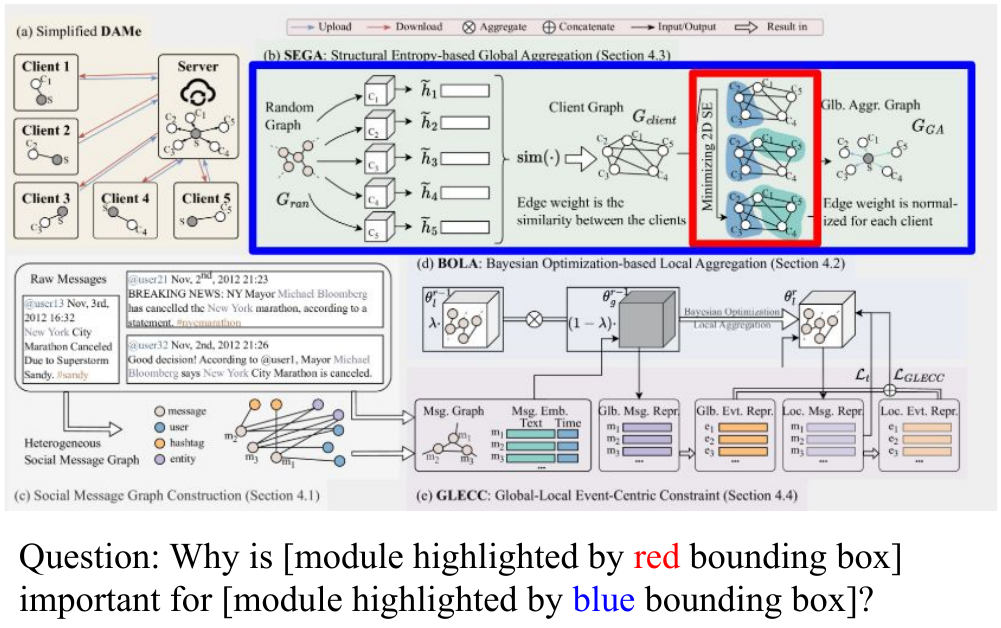}
    \caption{An example of a question posed within an information-seeking scenario related to design rationale.
    }
    \label{fig:missqa-example}
\end{figure}

\subsection{Question Annotation}
Annotators are given papers relevant to their fields to annotate questions following these three steps: 

\paragraph{Formulating Information-Seeking Question.} 
Annotators are initially given access to the abstract, introduction section, and schematic diagram of the paper. They are required to carefully review these materials to gain an overview of the study. 
They then identify the essential visual elements (\ie modules or procedures) within the diagram that are crucial to the research design and often lead to critical thinking or analysis regarding the study.
For annotating a single question, the annotation interface randomly assigns an information-seeking scenario (as described in \S\ref{sec:data-setup}) for the annotator to focus on. Based on this scenario, annotators formulate a question that incorporates the essential visual elements from the diagram.
Specifically, they are required to follow information-seeking scenarios, aiming to comprehend and navigate intricate information using the \figs for effective navigation to specific paper sections. The annotated questions are designed to require the models to first interpret the \fig, then navigate to the relevant sections of the paper for in-depth understanding and reasoning. 

\paragraph{Highlighting Relevant Visual Element.}
Recent studies have shown that visual content is unnecessary for many examples in current multimodal reasoning benchmarks, as questions can be directly answered from the textual part of the question due to the annotation bias or reasoning shortcut~\cite{yue2024mmmupro, chen2024mmstar, zhang2024mathverse}.
To alleviate this issue and ensure that the questions really necessitate interpretation of \fig within the paper, for each question, annotators apply one or two differently colored bounding boxes to highlight specific visual elements within the schematic diagram that are crucial for answering the questions.
In the question text, these areas are referred to as ``the module/process highlighted by [color] bounding box'', forcing models to interpret the figure rather than relying on textual clues within the question itself.
We also retain the original, unmasked questions within the dataset as a reference for future research.

\paragraph{Determining Question Answerability and Annotating Evidence.}
Annotators are initially provided only with the figure, abstract, and introduction sections of a paper to annotate questions. Under these constraints,  some questions are expected to be \emph{unanswerable} based on entire paper context. This is intentionally designed as, in real-world scenarios, researchers might ask information-seeking questions that cannot be answered based on the paper context. 
In such settings, the foundation models should acknowledge that the question is unanswerable instead of hallucinating responses~\cite{rajpurkar-etal-2018-know, asai-choi-2021-challenges}.
After completing the question annotation step, annotators are given access to the full paper to reassess the question's answerability. For questions deemed answerable, they must identify specific (sub)sections that provide the necessary information. These identified sources are subsequently used for verifying the annotated answers.

\subsection{Answer Annotation}\label{sec:answer-annotate}
For each question, a second annotator is assigned to annotate the answer through the following steps:
\paragraph{Preliminary Question Quality Check.}
The second annotator begins by evaluating the quality of the question to ensure it is clear, challenging, and truly requires the interpretation of the \fig as intended. The annotator is required to revise or discard those low-quality questions.

\paragraph{Annotating Evidence and Validating Alignment.}
Similar to the process in question annotation, the second annotator identifies and selects the relevant sections of the paper containing the necessary information to answer each question. Questions are labeled as ``unanswerable'' if no relevant information exists within the paper. 
The annotation interface compares the evidence annotated by both annotators. In cases of disagreement, the second annotator is provided with the first annotator’s inputs and asked to validate or revise their annotation. 
This iterative validation ensures consistency and quality across annotations.

\paragraph{Formulating Final Answer.}
Finally, the second annotator formulates a comprehensive and precise free-form answer based on the gathered evidence. Answers must avoid including any external information or knowledge beyond the given paper. Annotators are also encouraged to preserve original phrasing from the paper wherever appropriate to ensure accuracy and fidelity to the paper content.

\subsection{Data Quality Validation}
We implement a rigorous validation protocol to ensure the high quality of our annotated data. For each annotated QA example, a third annotator is assigned to validate the example based on several key aspects:
(1) the question must be meaningful, grammatically correct, and requires the interpretation of \figs for answering;
(2) The bounding boxes in the \figs must be accurately drawn and aligned with the question’s focus;
(3) the associated evidence is accurately and completely annotated;
(4) the answer is accurately annotated, directly linked to the provided evidence, and excludes any external information or knowledge not contained within the document.
Examples that fail to meet these standards are revised by the validators to ensure consistency and accuracy.

\input{tables/data_statistics}

\autoref{tab:basic-stats} presents the key data statistics of \ours. \ours include \qnum expert-annotated examples over \pnum papers.
We randomly divide the dataset into two subsets: \testmini for development validation and \test for standard evaluation. 

%% file: figure_tex/figure3.tex
\begin{table*}[!t]
\centering
\small
\begin{tabular}{p{0.36\textwidth}p{0.58\textwidth}}
\toprule
\textbf{Information-Seeking Scenarios} & \textbf{Example Question Provided in Appendix~\ref{app:example}} \\
\midrule
\textbf{Design Rationale}\newline Reasoning and motivations behind the structural and conceptual design choices &
\textcolor{white}{1}\newline What is the motivation behind designing the \emph{[module highlighted by red bounding box]} in the proposed framework?
\hfill(\autoref{fig:missqa-example} \& Appendix ~\ref{app:e1})
\\

\midrule 
\textbf{Implementation Details}\newline Technical details of how the specified modules or procedures are implemented &
\textcolor{white}{1}\newline 
How does the \emph{[module highlighted by red bounding box]} interact with the \emph{[module highlighted by green bounding box]}? \hfill(Appendix ~\ref{app:e2})
\\

\midrule 
\textbf{Literature Background}\newline Relationship between the specified modules and relevant prior research or methodologies &
\textcolor{white}{1}\newline
Which prior work does the \emph{[module highlighted by red bounding box]} build on?
\hfill(Appendix ~\ref{app:e3})
\\

\midrule 
\textbf{Experimental Results}\newline Key findings related to the specified modules or procedures based on experimental results &
\textcolor{white}{1}\newline
How does the \emph{[module highlighted by red bounding box]} contribute to the overall framework based on the results?
\hfill (Appendix ~\ref{app:e4})
\\

\midrule 
\textbf{Others}\newline Additional aspects such as limitations, alternative approaches, and ethical considerations. &
\textcolor{white}{1}\newline
What are the limitations of the [module highlighted by the red bounding box] as discussed in the paper?
\hfill (Appendix ~\ref{app:e5})
\\
\bottomrule
\end{tabular}
\caption{
Definition of the investigated information-seeking scenarios in \ours.
}
\label{tab:data-type}
\end{table*}

%% file: tables/data_statistics.tex





\begin{table}[t!]
\centering 
\renewcommand{\arraystretch}{1.2}
\small
\addtolength{\tabcolsep}{-0.3em}
\begin{tabular}{lrr}
\hline
\toprule
\textbf{Property} (\texttt{Median/Avg})  & \textbf{Test} \\
\midrule
$\#$ Scientific Papers &  465  \\
Schematic Diagram Caption Len.  & 57 / 67  \\ 

Paper Length    &  4,368 / 4,565    \\ 

Question Length   & 14.1 / 13.8   \\


Answer Length  &    59.0/ 60.3             \\
\midrule
$\#$ Unanswerable Questions  & 398   \\
$\#$ Answerable Questions  & 1102    \\
$\#$ QA Examples  & 1,500  \\
\bottomrule
\hline
\end{tabular}

\caption{Basic statistics of \ours. Answer length statistics exclude unanswerable questions.}
\label{tab:basic-stats}
\end{table}

%% file: main/4-2-experiment.tex
\section{Experiment}
This section presents the experiment setup, followed by a detailed analysis of the results.

\subsection{Experiment Setup}
\paragraph{Automated Evaluation System.}
We use accuracy as the primary metric to evaluate model performance on \ours. Following recent benchmark studies~\cite{lu2024mathvista, li-etal-2024-m3sciqa}, we adopt the LLM-as-Judge framework with GPT-4.1 as the base evaluator. Specifically, the model is prompted to assign an accuracy score of 0, 0.5, or 1 by comparing the generated response with the ground-truth answer.

\paragraph{Human Expert Performance Measurment.}
To estimate human expert-level performance on \ours, we randomly selected 50 examples, comprising 10 from each subset of the \testmini set. Two PhD candidates with specializations in NLP and CV, respectively, were tasked with independently solving these questions within 5 hours. Their responses were evaluated using our developed automated evaluation protocol. They achieve an average accuracy of 89.0\% (\autoref{tab:main-results}).

\paragraph{Evaluated Multimodal Foundation Models.}
\ours requires models to interpret the schematic diagrams, therefore, we only evaluate foundation models with vision capabilities in this study. Specifically, we examine the performance of \nmodel \lmm across two distinct categories on \ours: 
(1) \textbf{proprietary models}, including 
OpenAI o4-mini~\cite{openai2025o4mini}, GPT-4o, GPT-4.1 and GPT-4.1-mini ~\cite{openai2024gpt4o,openai2024gpt4-1} and  
Gemini-2.5-Flash~\cite{geminiteam2024gemini}, 
and (2) \textbf{open-source models}, including 
Qwen2-VL and  Qwen2.5-VL~\cite{wang2024qwen2vl},
InternVL-2, 2.5 and 3~\cite{chen2023internvl, chen2024far,chen2024expanding},
Pixtral~\cite{pixtral},
Mistral-Small-3.1~\cite{mistral2025small31},
and Phi-3.5-Vision and and Phi-4-Multimodal~\cite{abdin2024phi3,microsoft2025phi4}.
\autoref{tab:model-info} in Appendix presents the details of each evaluated model in this study. 
\autoref{fig:cot_prompt} in Appendix presents the prompt employed in our main experiments, which is adapted from M3SciQA~\cite{li-etal-2024-m3sciqa}.

\input{tables/main_results}
\subsection{Main Findings}\label{sec:main-result}
A substantial performance gap remains between human experts and state-of-the-art open-source models. For example, Qwen2.5-VL-72B, the highest-performing open-source model to date, achieves an accuracy of only 61.6\%, significantly trailing behind the 89.0\% accuracy attained by human experts. While proprietary models such as o4-mini and GPT-4.1 have made notable strides in narrowing this gap, open-source models still lag in overall performance.
Nevertheless, recent developments in open-source models show promising progress. Notable improvements can be observed within the same model families: Qwen2.5-VL-72B outperforms its predecessor Qwen2-VL-72B by 7.4\%, and InternVL3-38B surpasses InternVL2.5-38B by 8.9\%. These advancements underscore the rapid evolution of open-source technologies, driven by ongoing innovation and community collaboration.

Interestingly, we find that most models, except for Gemini-2.5-Flash and certain lower-performing models like the Phi series, continue to struggle with unanswerable questions. These models often exhibit overconfidence, attempting to provide answers even when the correct response is that no answer exists within the given paper context. In contrast, human experts display greater robustness and discernment in such scenarios.

\input{figure_tex/confusion}
\input{appendix/error_examples/error_in_main}
\subsection{Analysis on Unanswerable Question.}\label{sec:unanswer-analysis}
Reflecting real-world information-seeking scenarios where researchers may encounter questions that cannot be answered based on the context of the papers they are reading, 
26.5\% of the questions in \ours are designed to be unanswerable (\S\ref{sec:answer-annotate}). 
This subset aims to evaluate models' ability to handle uncertain queries and appropriately respond by indicating the lack of sufficient information. \autoref{fig:confusion} illustrates the confusion matrices for answerable and unanswerable questions, comparing the responses of two top-performing proprietary models and two open-source models against the ground-truth answers on the test set.

\subsection{Error Analysis and Case Study.}
To gain deeper insights into the strengths and limitations of the top-performing open-source multimodal foundation models on our task, we conducted a thorough error analysis. This analysis examines 100 randomly sampled error cases, with 50 cases each from Qwen2-VL-72B and InternVL2.5-38B, derived from the \testmini set. 
We identify the following five common error types: 

\noindent \textbf{(1) Failure to interpret and contextualize schematic diagrams}: The model struggles to decode the structure and meaning of the schematic diagram, failing to identify highlighted visual elements or associate them with relevant paper context. Examples are presented in Appendix~\ref{app:err1}.
\textbf{(2) Inability to retrieve relevant context}: The model is unable to effectively locate or extract the sections of the paper most relevant to addressing the question, leading to incomplete or irrelevant answers. Examples are presented in Appendix~\ref{app:err2}.
\textbf{(3) Reasoning error}: The model successfully identifies the relevant visual and textual information needed to answer the question but fails to reason correctly with this information, leading to incorrect response. Examples are presented in \autoref{fig:error_in_main} and Appendix~\ref{app:err3}.
\textbf{(4) Overconfident response to unanswerable questions}: While the model successfully interprets the schematic diagram and visual elements, it generates responses even when the provided
information does not suffice to answer the question. Examples are presented in Appendix~\ref{app:err4}.
\textbf{(5) Overreliance on visual elements}: The model disproportionately focuses on the schematic diagram, neglecting the given paper context. They rely on isolated visual interpretation with parametric knowledge to answer the question. Examples are presented in Appendix~\ref{app:err5}.
Other errors include generating a response that exceeds the 1024-token output limit, and refusing to answer due to safety alignment.
Our findings suggest that the primary sources of errors are the models' difficulties in interpreting and contextualizing schematic diagrams and retrieving relevant contextual information from papers to construct accurate answers.

%% file: tables/main_results.tex
\begin{table*}[t!]
\centering
\renewcommand{\arraystretch}{1.05}
\small
\begin{tabular}{l*{6}{>{\centering\arraybackslash}p{1.2cm}}>{\columncolor[gray]{0.95}\centering\arraybackslash}p{1cm}>{\columncolor[gray]{0.95}\centering\arraybackslash}p{1cm}}
\hline
\toprule
\textbf{Model} 
& 
\begin{tabular}[c]{@{}c@{}}
\textbf{Design}  \\
\textbf{Rationale}
\end{tabular} 
& 
\begin{tabular}[c]{@{}c@{}}
\textbf{Implem.}  \\
\textbf{Details}
\end{tabular} 
& 
\begin{tabular}[c]{@{}c@{}}
\textbf{Literature}  \\
\textbf{Bkgd.}
\end{tabular} 
& 
\begin{tabular}[c]{@{}c@{}}
\textbf{Experiment}  \\
\textbf{Results}
\end{tabular} 
& \textbf{Other}
&
\begin{tabular}[c]{@{}c@{}}
\textbf{Unaswer-}  \\
\textbf{able}
\end{tabular} 
& \textbf{Testmini}
& \textbf{Test}
\\
 & (207) & (212) & (201) & (193) & (187) & (225) & (500) & (1,000) \\

\midrule
\multicolumn{7}{c}{\emph{Baselines}} \\
\noalign{\vskip 1ex}
Human Expert
& 90.0 & 85.0 & 95.0 & 95.0 & 80.0 & 85.0 & \multicolumn{2}{c}{89.0}\\

\midrule
\multicolumn{8}{c}{\emph{Proprietary Multimodal Foundation Models}}\\
\noalign{\vskip 1ex}
o4-mini & \underline{88.2} & \textbf{84.9} & \textbf{84.8} & \textbf{86.0} & 44.9 & 33.8 &\underline{72.1}& \textbf{78.3} \\
GPT-4.1 & 85.0 & \underline{76.4} & \underline{77.9} & \underline{82.1} & \underline{67.1} & \underline{60.0} &\textbf{73.9}& \underline{77.8} \\
GPT-4.1-mini & \textbf{88.4} & 74.8 & 75.1 & 77.2 & 53.2 & 42.2 &67.8& 74.1 \\
Gemini-2.5-Flash & 72.2 & 63.4 & 57.2 & 57.8 & \textbf{87.2} & \textbf{88.0 }&60.3&67.3 \\
GPT-4o & 73.9 & 67.5 & 67.2 & 65.8 & 38.5 & 34.7 &57.4& 63.0 \\

\midrule
\multicolumn{8}{c}{\emph{Open-Source Multimodal Foundation Models}}\\
Qwen2.5-VL-72B & \textbf{74.2} & 62.3 & \textbf{64.4} & \textbf{64.0} & 41.4 & 33.3 &\textbf{54.9}& \textbf{61.6} \\
InternVL3-38B & \underline{72.7} & \underline{62.7} & \underline{60.9} & \underline{61.9} & 41.7 & 34.2 &52.0& \underline{60.4} \\
Mistral-Small-3.1-24B & 64.0 & \textbf{63.4} & 57.5 & 54.7 & 45.5 & 47.1 &\underline{52.4}& 57.3 \\
Qwen2-VL-72B & 62.3 & 55.2 & 50.7 & 52.8 & 49.5 & 46.2 &49.8& 54.2 \\
InternVL2.5-38B & 56.5 & 49.5 & 50.2 & 51.6 & 50.3 & 52.0 &46.9& 51.6 \\
InternVL3-8B & 58.2 & 47.2 & 55.2 & 53.6 & 43.0 & 36.9 &48.3& 51.5 \\
Pixtral-12b & 52.7 & 55.4 & 49.8 & 48.2 & 41.7 & 43.1 &45.1& 49.8 \\
Qwen2.5-VL-7B & 45.9 & 40.8 & 38.1 & 39.9 & 46.0 & 37.8 &33.8& 42.1 \\
Qwen2-VL-7B& 37.7 & 29.2 & 34.3 & 37.6 & 52.7 & 64.9 &32.8& 38.0 \\
InternVL2.5-8B & 37.2 & 31.4 & 33.8 & 32.4 & 44.4 & 44.0 &37.6& 35.7 \\
Phi-4-Multimodal & 14.3 & 23.8 & 32.1 & 21.8 & \textbf{73.5} & \textbf{93.3} &36.7& 32.4 \\
InternVL2-8B & 30.4 & 26.2 & 27.4 & 30.3 & 46.8 & 52.4 &32.6& 31.9 \\
Phi-3.5-Vision & 18.8 & 22.6 & 29.6 & 23.3 & \underline{59.9} & \underline{81.3} &35.1& 30.3 \\

\bottomrule
\hline
\end{tabular}
\caption{
Model performance on the \ours testmini and test splits, with a detailed breakdown across various subsets of the test split. Test set accuracy serves as the ranking indicator. \textbf{Bold} and \underline{underlined} numbers indicate the best and second-best performance in each category, respectively. 
}
\label{tab:main-results}
\end{table*}

%% file: figure_tex/confusion.tex
\begin{figure}[!t]
    \centering
    \includegraphics[width = 0.9\linewidth]{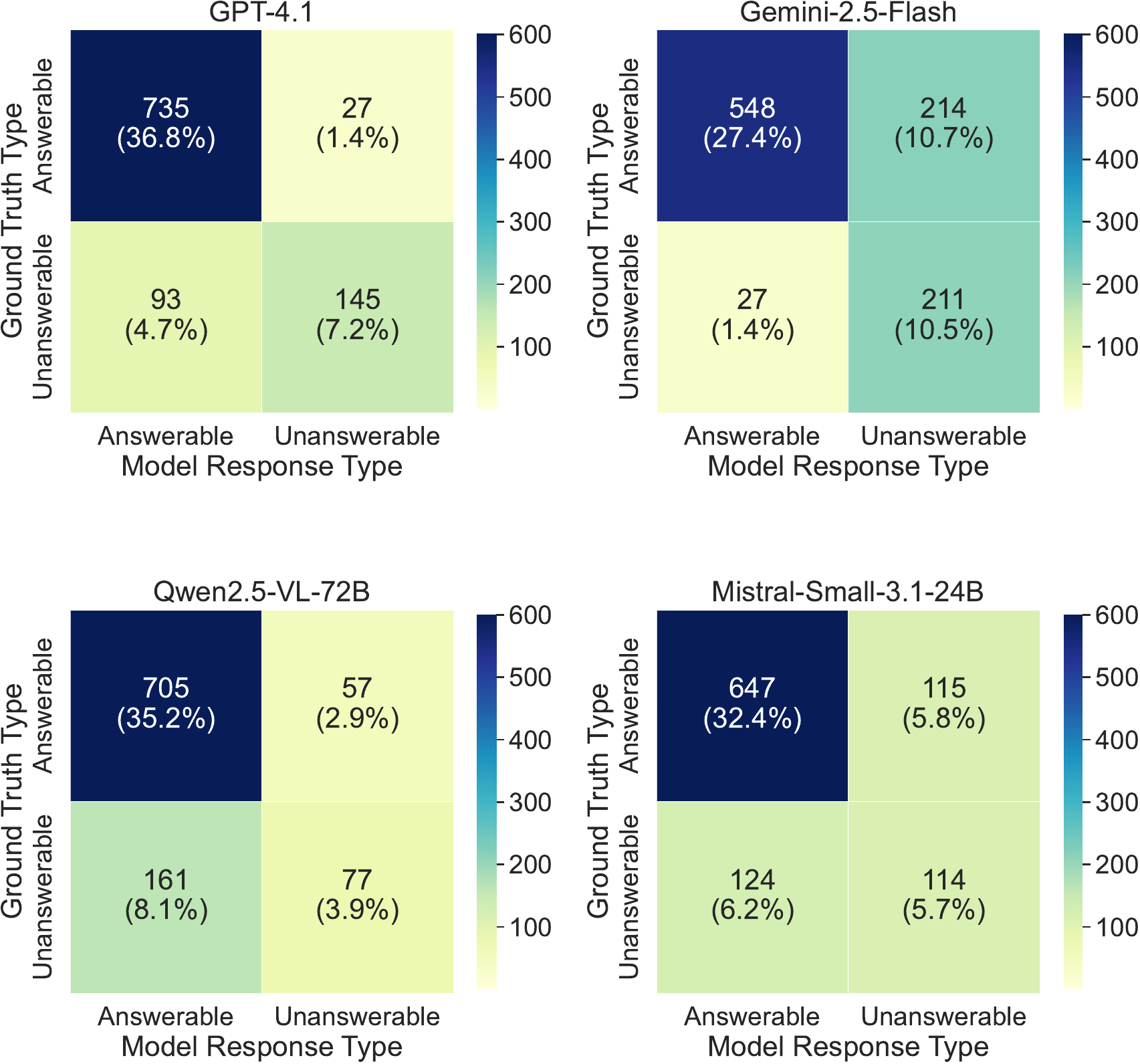}
    \caption{Confusion matrices comparing the model response types (whether answerable) to the ground truth on the test set.
    }
    \label{fig:confusion}
\end{figure}

%% file: appendix/error_examples/error_in_main.tex
\begin{figure*}[!t]               
    \centering
\begin{blueBox}[Design Rationale]
\begin{center}
    \includegraphics[width=0.8\linewidth]{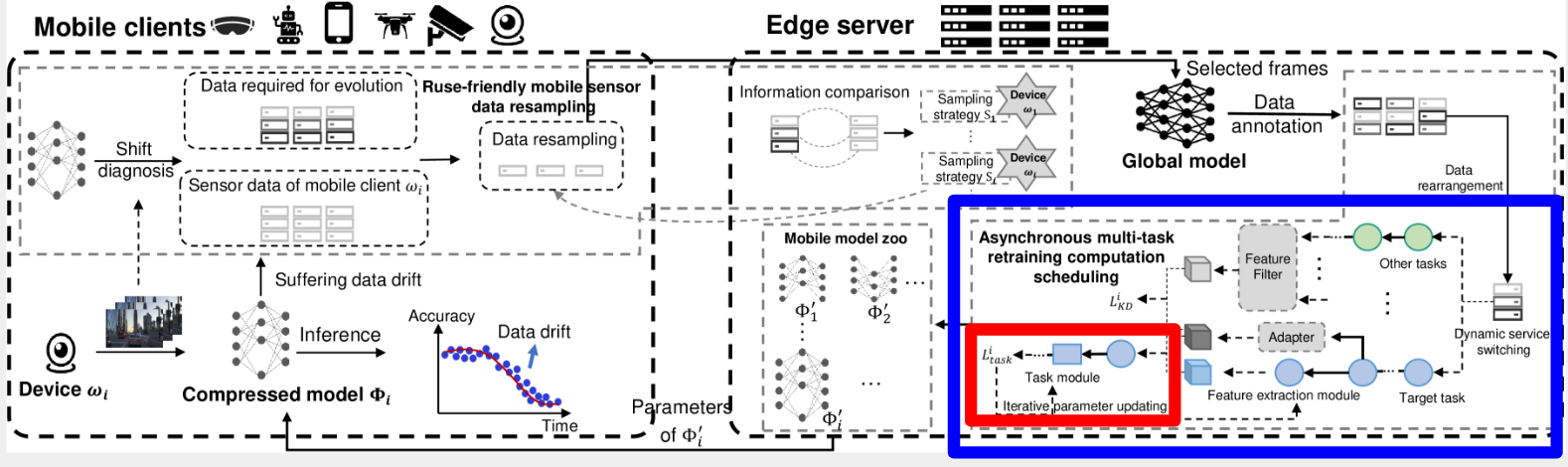}
\end{center}

\small\textbf{Paper Title:}
\newline
AdaBridge: Dynamic Data and Computation Reuse for Efficient Multi-task DNN Co-evolution in Edge Systems

\textbf{Figure Caption:}
\newline
Illustration of AdaBridge s system workflow.

\textbf{Question:}
\newline
What is the importance of [module highlighted by red bounding box] in [module highlighted by blue bounding box]?

\textbf{Ground-truth Answer:} 
\newline
The importance of iterative parameter updating in retraining scheduling is to optimize the limited representation capacity of compressed mobile DNNs by incorporating dynamic switching and incremental parameter updating.

\small \textbf{Relevant Paper Context:} 
\newline
Section 2: \hspace{3mm}AdaBridge Design:
\newline
\emph{``It asynchronously reuses learned features from different subtasks and incorporates dynamic switching and incremental parameter updating to optimize the limited representation capacity of compressed mobile DNNs''}

\end{blueBox}

\begin{wronganswer}
\small \textbf{Qwen2-VL-72B:}
The module highlighted by the red bounding box is the ``Task module'' within the "Asynchronous multi-task retraining computation scheduling" module (highlighted by the blue bounding box). The Task module is important because it is responsible for performing the specific tasks that the system is designed to accomplish. The iterative parameter updating within the Task module allows the system to adapt and improve its performance over time, which is crucial for maintaining accuracy and efficiency in the face of changing data and evolving tasks.

\end{wronganswer}

\begin{correctanswer}
\small\textbf{Error Analysis:}\\
The Qwen2-VL-72B's answer misses the core technical importance of iterative parameter updating in terms of improving dynamic adaptability and optimizing limited representation capacity. It does not delve into how this specific process aids in mitigating challenges like data drift or evolving tasks.

\end{correctanswer}
\caption{An example of model reasoning error.}
\label{fig:error_in_main}
\end{figure*}

%% file: main/5-conclusion.tex
\section{Conclusion}

This paper presents \ours, the first benchmark specifically designed to evaluate multimodal foundation models in their ability to interpret schematic diagrams within scientific literature.
We perform a comprehensive evaluation of \nmodel frontier multimodal foundation models, uncovering a significant performance disparity between human experts and current models.
Through an in-depth assessment, we identify the weaknesses of cutting-edge foundation models on \ours, offering valuable insights to guide future advancements in the field.

\section*{Acknowledgement}
We are grateful to NVIDIA Academic Grant Program for providing computing resources and Together AI for granting LLM API credits.

%% file: main/limitations.tex
\section*{Limitations}
While our benchmark aims to reflect real-world research scenarios, it predominantly focuses on AI-related scientific papers from arXiv, as our expert annotators specialize in various subfields within the AI domain. Extending this work to other scientific disciplines, such as biology and medicine, would enhance its generalizability and applicability across broader research areas.
Moreover, we examined the available documentation for each evaluated model, but unfortunately, most models do not disclose detailed information about their pretraining data. This lack of transparency prevents us from definitively determining whether scientific content was included in their training pipeline and, in turn, limits our ability to correlate performance directly to domain-specific training.

%% file: appendix/a1.tex
\clearpage
\onecolumn
\section{\ours Benchmark}
\subsection{Annotator Biographies}\label{sec:annotator-info}
\input{appendix/annotator_info}

\subsection{Annotation Interface}\label{app:interface}
\begin{figure*}[h]
    \centering
    \includegraphics[width = 0.98\linewidth]{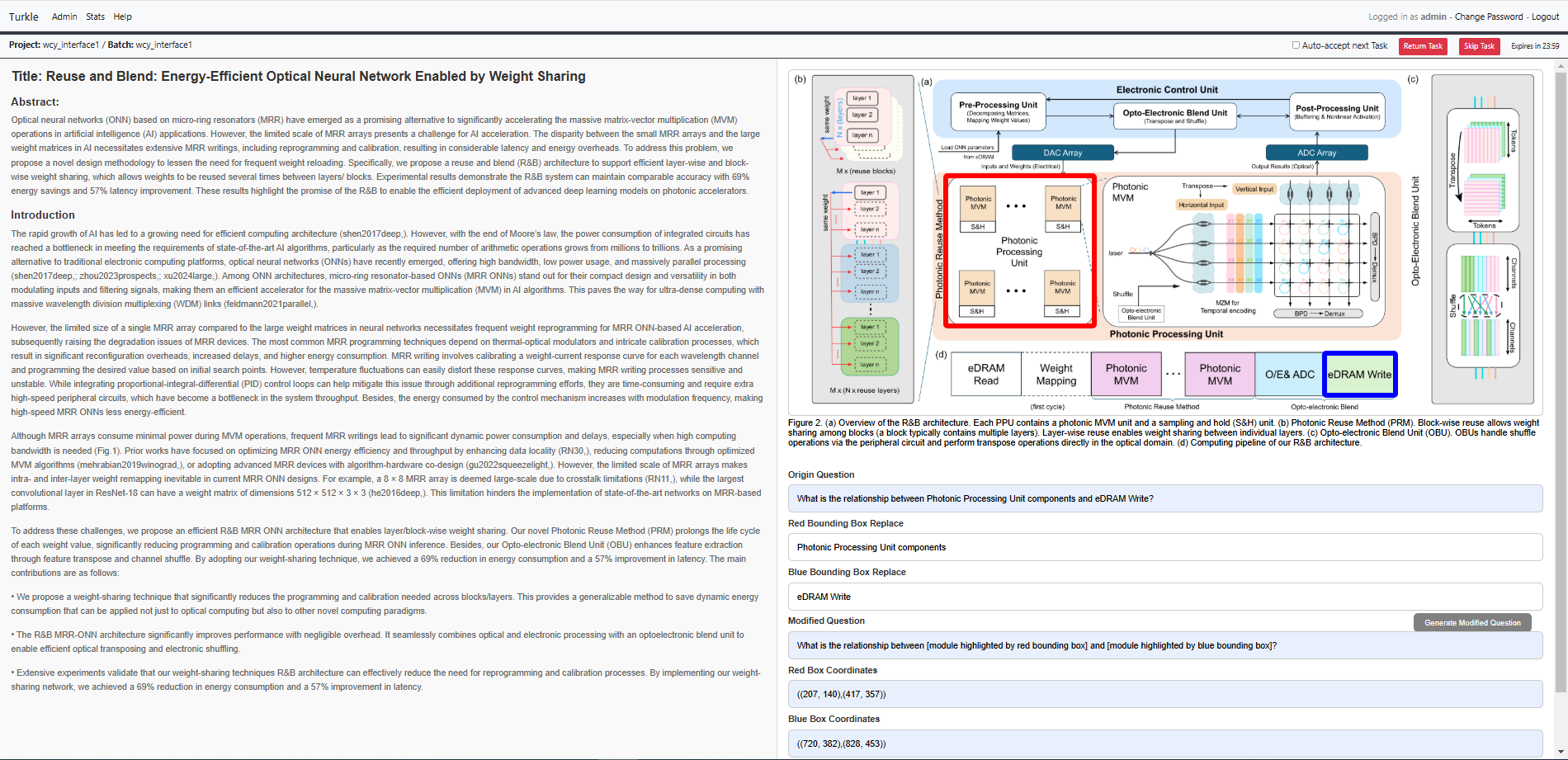}
    \caption{The interface used by the first annotator for question annotation, where annotators highlight question-relevant visual elements in the schematic diagram using colored bounding boxes. The positions of the bounding boxes are automatically captured and recorded.}
\end{figure*}

\begin{figure*}[h]
    \centering
    \includegraphics[width = 0.98\linewidth]{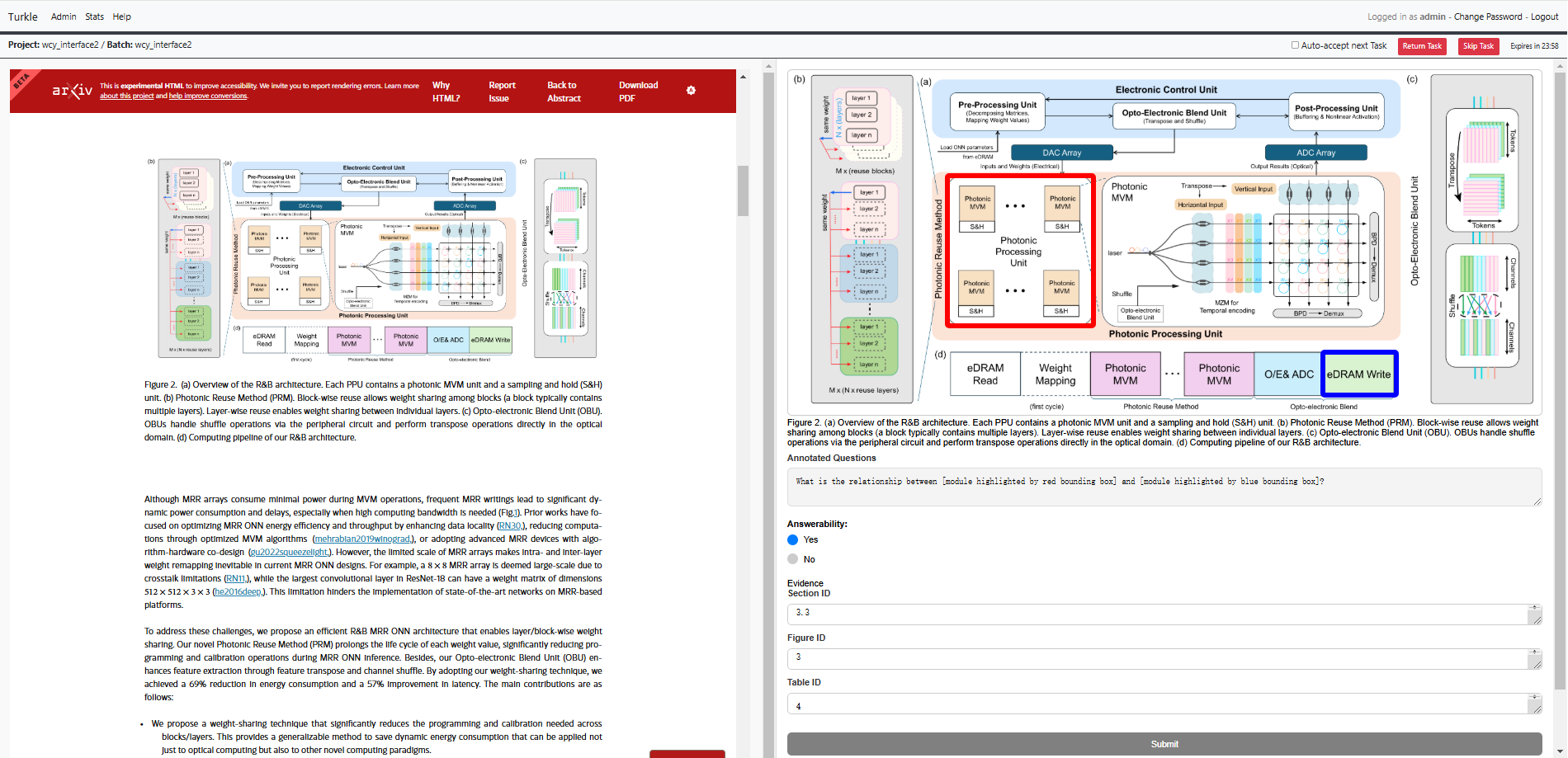}
    \caption{The interface used by the first annotator to evaluate question answerability and identify supporting evidence following question annotation.}
\end{figure*}

\begin{figure*}[h]
    \centering
    \includegraphics[width = 0.98\linewidth]{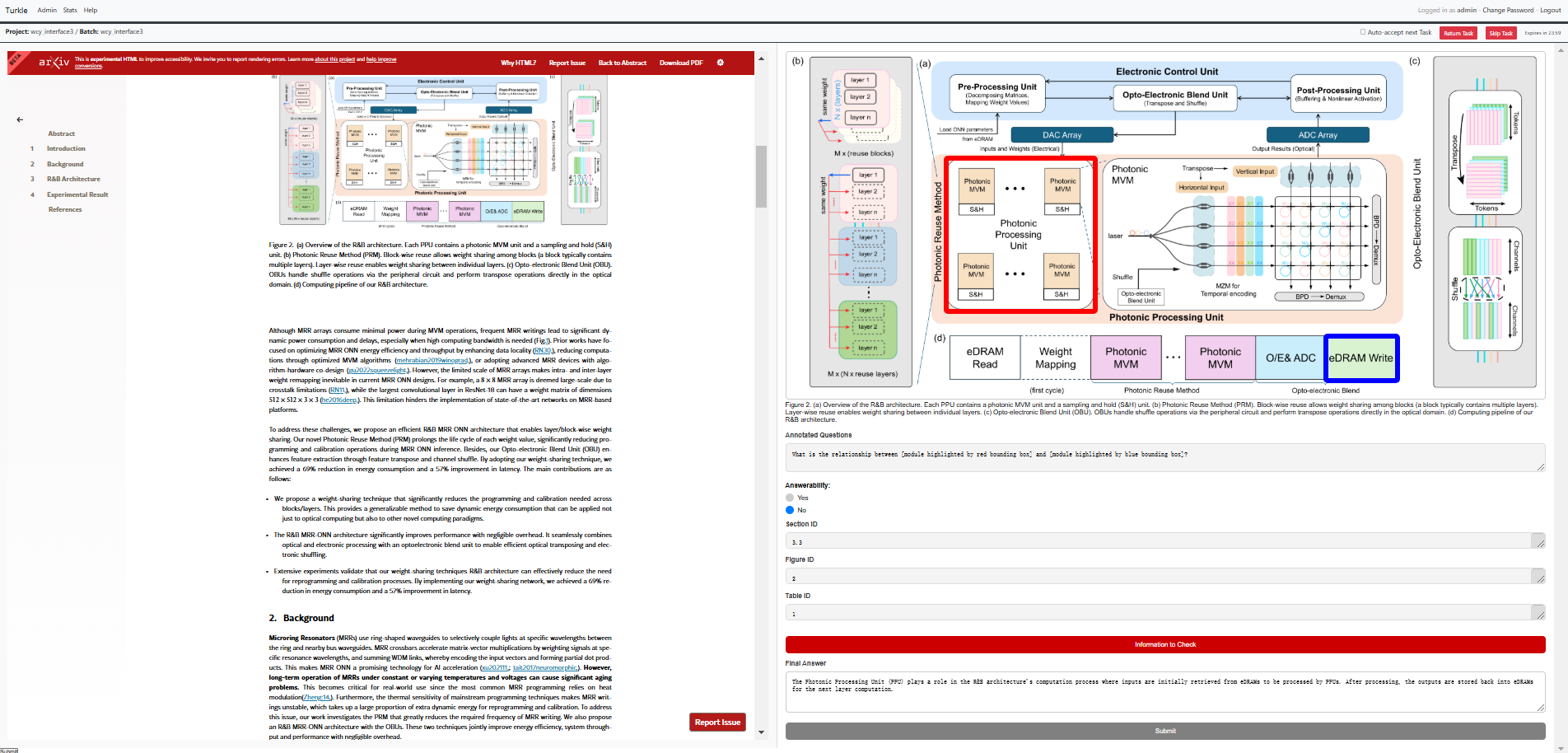}
    \caption{The interface used by the second annotator for answer annotation. The annotator begins by entering the supporting evidence and then compares it with the annotation provided by the first annotator. This process enhances the accuracy and consistency of the answer annotations.}
\end{figure*}
\clearpage

\section{Dataset Examples of Different Information-Seeking Scenarios}\label{app:example}
\input{appendix/scenario_examples/Design_Rationale}
\input{appendix/scenario_examples/Design_Rationale2}
\input{appendix/scenario_examples/Implementation_Details}
\input{appendix/scenario_examples/Implementation_Details2}
\input{appendix/scenario_examples/Literature_Background}
\input{appendix/scenario_examples/Literature_Background2}
\input{appendix/scenario_examples/Experimental_Results}
\input{appendix/scenario_examples/Experimental_Results2}
\input{appendix/scenario_examples/Others}
\input{appendix/scenario_examples/Others2}
\clearpage

\section{Experiment Setup}
\input{appendix/model_configuration}
\input{figure_tex/cot_prompt}

\onecolumn
\section{Error Analysis and Case Study}\label{app:err}
\input{appendix/error_examples/err1}
\clearpage
\input{appendix/error_examples/err2}
\clearpage
\input{appendix/error_examples/err3}
\clearpage
\input{appendix/error_examples/err4}
\clearpage
\input{appendix/error_examples/err5}

%% file: appendix/annotator_info.tex
\begin{table*}[h]
\centering
\small
\begin{tabular}{lllccc}
\toprule
 ID &          Position &         Field & Publications & Data Annotation & Data Validation \\
\midrule
  1 & Research Scientist &             ML &  > 10 & \cmark &        \\
  2 &           Postdoc &             CV &  > 10 & \cmark & \cmark \\
  3 & Research Scientist &             ML &  > 10 &   \cmark     & \cmark \\
  4 &         PhD Student &            NLP &  > 10 & \cmark & \cmark \\
  5 &         PhD Student &            NLP &  5-10 & \cmark &        \\
  6 &         PhD Student &             CV &   3-5 & \cmark & \cmark \\
  7 &         PhD Student &        CV/NLP &  5-10 & \cmark &        \\
  8 &           Postdoc &            NLP &  > 10 &   \cmark     & \cmark \\
  9 &         PhD Student &        CV/NLP &   3-5 & \cmark &        \\
 10 &         PhD Student &             DM &  5-10 & \cmark &        \\
 11 & Research Scientist &            NLP &  > 10 &   \cmark     & \cmark \\
 12 &         PhD Student &             ML &  > 10 &        & \cmark \\
 13 &         PhD Student &        IR/NLP &  5-10 & \cmark &        \\
 14 &           Postdoc &            NLP &   3-5 & \cmark & \cmark \\
 15 &         PhD Student &        IR/NLP &  5-10 &   \cmark     & \cmark \\
 16 &         PhD Student &             ML &   3-5 & \cmark &        \\
\bottomrule
\end{tabular}

\caption{Details of annotators involved in dataset construction. \ours is annotated by experts in AI domains, ensuring both the accuracy of the benchmark and the reliability of the human evaluation.}
\label{tab:candidate_profiles}
\end{table*}

%% file: appendix/scenario_examples/Design_Rationale.tex
\subsection{Design Rationale Diagram example}\label{app:e1}
\begin{blueBox}[Design Rationale]
\begin{center}
    \includegraphics[width=0.8\linewidth]{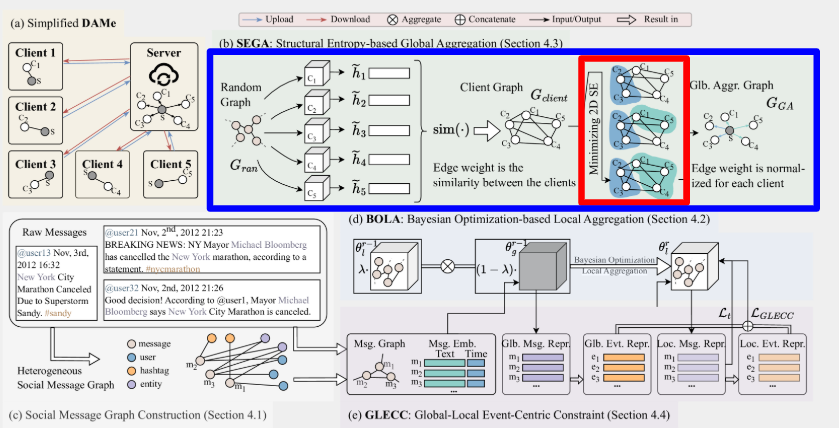}
\end{center}

\small \textbf{Paper Title:}
\newline
DAMe: Personalized Federated Social Event Detection with Dual Aggregation Mechanism
\newline

\textbf{Figure Caption:}
\newline
The overall framework of DAMe.
\newline

\textbf{Question:}
\newline
Why is [module highlighted by red bounding box] important for [module highlighted by blue bounding box]?
\newline

\textbf{Ground-truth Answer:} 
\newline
Minimizing 2D structural entropy is important for global aggregation because it enables the creation of a client graph that captures the internal similarities among clients, which is used to finalize the Global Aggregation strategy, maximizing the external knowledge available to each client while considering client heterogeneity and individual characteristics.
\newline

\small \textbf{Relevant Paper Context:} 
\newline
Section 4.3 \hspace{3mm} Global Aggregation via 2D Structural Entropy Minimization:
\newline
\emph{``Under the federated framework described in Section 3.1, personalized global aggregation aims to provide clients with maximum external information by producing global models that can benefit individual clients more. The server needs an aggregation strategy that considers client heterogeneity and individual characteristics to maximize external knowledge for all clients. To achieve this objective, we construct a client graph Gclient based on clients’ similarity. By minimizing the two-dimensional Structural Entropy (2DSE) of Gclient, a graph capturing the internal similarities among clients is obtained, finalizing the Global Aggregation strategy for each client (SEGA).Gclient is an undirected, fully connected, weighted graph consisting of K nodes corresponding to K clients, with their similarities as edge weights. The similarity between client models can be estimated by providing them with the same input and measuring the similarity between their respective outputs. On this basis, the server first generates a random graph Grandom as input to all client models. With graph pooling, the server obtains different client models’ representations of the same graph''}

\end{blueBox}
\clearpage

%% file: appendix/scenario_examples/Design_Rationale2.tex
\begin{blueBox}[Design Rationale]
\begin{center}
    \includegraphics[width=0.8\linewidth]{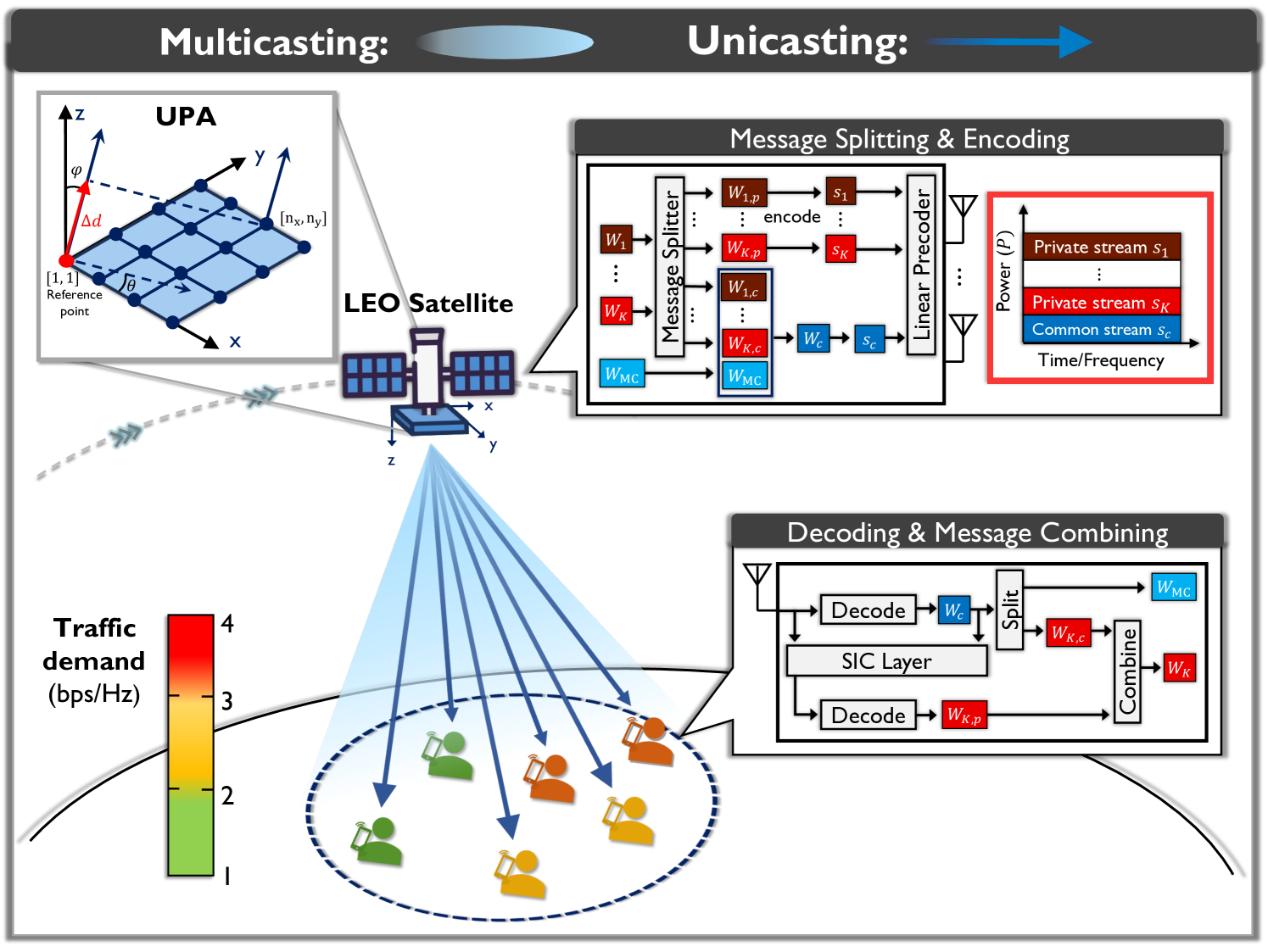}
\end{center}

\small \textbf{Paper Title:}
\newline
Rate-Splitting for Joint Unicast and Multicast Transmission in LEO Satellite Networks with Non-Uniform Traffic Demand
\newline

\textbf{Figure Caption:}
\newline
System model of the proposed RSMA-based NOUM transmission.
\newline

\textbf{Question:}
\newline
What is the motivation behind combining the content of [module highlighted by red bounding box] in this framework?
\newline

\textbf{Ground-truth Answer:} 
\newline
The motivation behind combining common and private streams in this framework is to effectively minimize the unused and unmet rates by allocating usable power according to traffic requirements, manage interference between unicast and multicast streams, and ensure robustness against imperfect channel state information.
\newline

\small \textbf{Relevant Paper Context:} 
\newline
Section 1.2 \hspace{3mm} Motivations and Contributions:
\newline
\emph{``We propose an RSMA-based RM framework that minimizes the difference between traffic demands and actual offered rates for both unicast and multicast messages. By flexibly allocating the usable power into the common and private streams according to the traffic requirements, the unused/unmet rates are effectively minimized. To cope with the challenge of obtaining instantaneous CSIT at LEO satellites, we leverage the statistical and geometrical information of satellite-to-user channels, which vary comparably slower, in the RSMA precoder design.''}

\end{blueBox}
\clearpage

%% file: appendix/scenario_examples/Implementation_Details.tex
\subsection{Implementation Details Diagram example}\label{app:e2}
\begin{blueBox}[Implementation Details]
\begin{center}
    \includegraphics[width=0.8\linewidth]{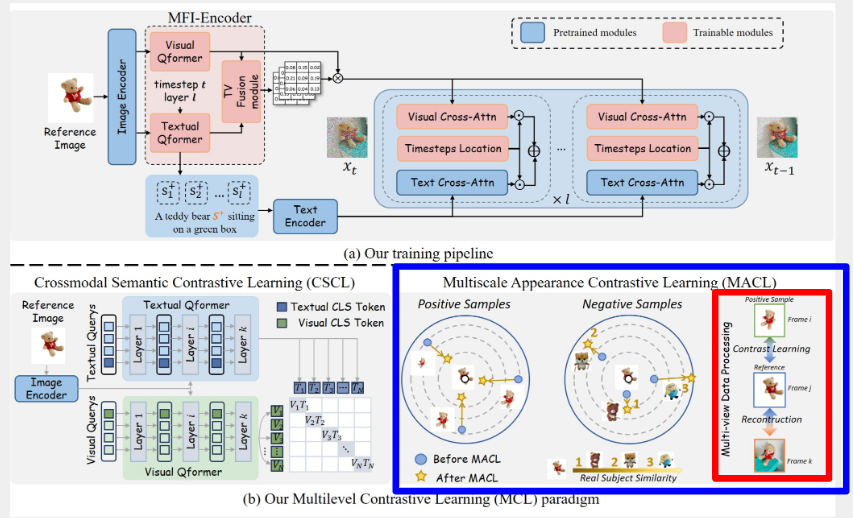}
\end{center}

\small\textbf{Paper Title:}
\newline
CustomContrast: A Multilevel Contrastive Perspective For Subject-Driven Text-to-Image Customization
\newline

\textbf{Figure Caption:}
\newline
Overview of the proposed CustomContrast. (a) Training pipeline. The consistency between textual and visual features is accurately learned by the MFI-Encoder, which includes a Textual-Visual (TV) Fusion module to enhance feature consistency from visual and textual Qformers. (b) The MCL paradigm includes CSCL, aligning high-level semantics by contrasting visual and textual embeddings via CLS tokens, and MACL,
which is applied to text embeddings from different cross-attention layers. MACL decouples redundant subject features by aligning positive samples (segmented images of the same subject from various views, positions, and sizes), while preserving relative distances by contrasting with other subjects.
\newline

\textbf{Question:}
\newline
How does [module highlighted by blue bounding box] achieve real subject similarity using [module highlighted by red bounding box]?
\newline

\textbf{Ground-truth Answer:} 
\newline
MACL achieves real subject similarity using multiview data by ensuring that the distance relationships between multiscale features are consistent with those of real subjects. This is done by maintaining intra-consistency, where features of the same subject with different situations are as close as possible, and inter-distinctiveness, where the distances between different samples' features match those between real subjects. MACL preserves the multi-scale similarity structure, ensuring that the similarities of learned features are positively correlated with those of real subjects.
\newline

\small \textbf{Relevant Paper Context:} 
\newline
Section 3.3.2\hspace{3mm}Multiscale Appearance Contrastive Learning:
\newline
\emph{``The key idea of Multi-scale Appearance Similarity Contrastive Learning (MACL) is to ensure that the distance relationships between multiscale features are consistent with those of real subjects. This means the features of the same subject with different situations should be as close as possible (intra-consistency), while the distances between different samples’ features should match those between real subjects (inter-distinctiveness). As shown in Fig. 2(b)(right), we achieve intra-consistency by pulling positive samples of the reference subject closer, and inter-distinctiveness by introducing scaling factors to align the feature distances with negative samples to real subject distances. In this section, we will introduce the S+Space and MACL in the S+Space. As shown in Fig. 2(b) (right), We select frames different from the reference images as MACL positive samples. By aligning images of the same subject, CustomContrast effectively decouples irrelevant features of the subject. The processing details of positive samples are in Appendix B.''}

\end{blueBox}
\clearpage

%% file: appendix/scenario_examples/Implementation_Details2.tex
\begin{blueBox}[Implementation Details]
\begin{center}
    \includegraphics[width=0.8\linewidth]{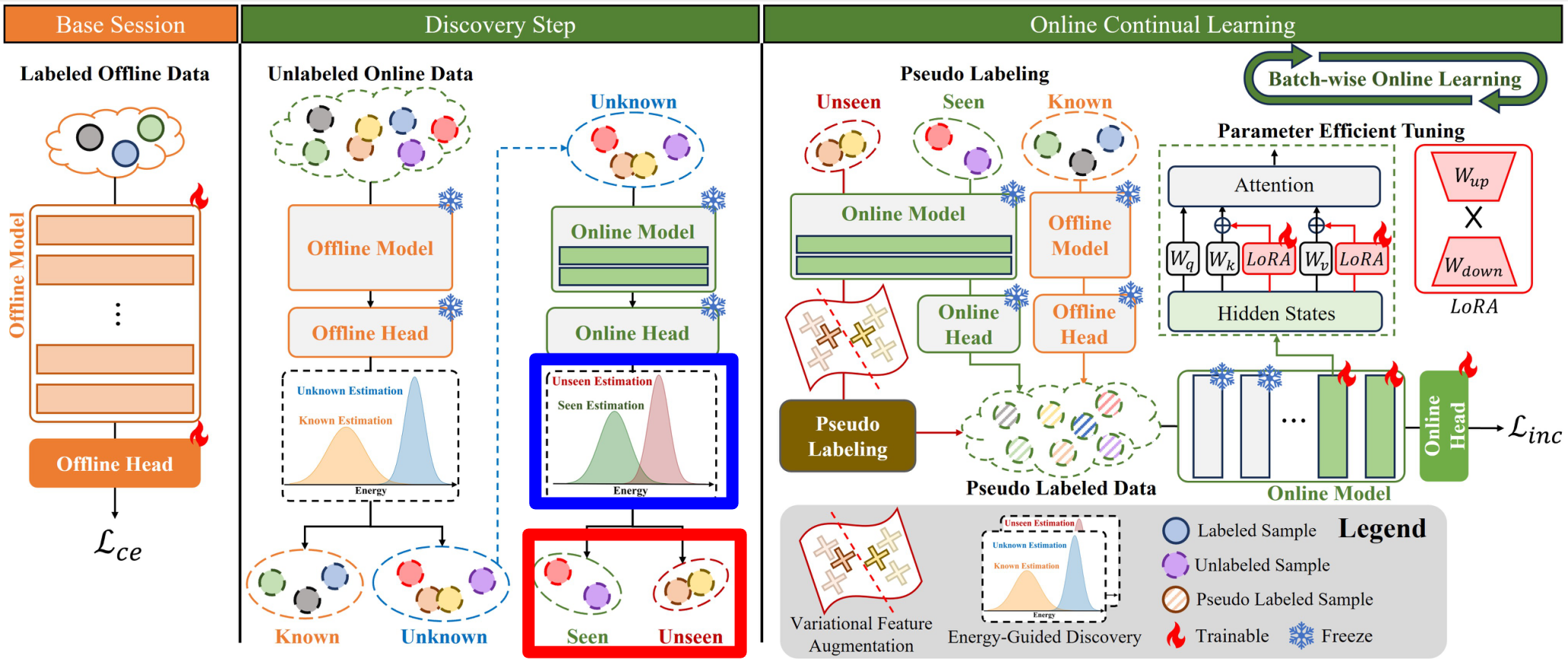}
\end{center}

\small\textbf{Paper Title:}
\newline
Online Continuous Generalized Category Discovery
\newline

\textbf{Figure Caption:}
\newline
Overall process of the proposed DEAN framework. The energy-guided discovery splits unlabeled data into known, seen, and unseen data for better novel category discovery, while variance-based feature augmentation enhances the clustering of unseen data. Lce facilitates better discriminative learning in the online continual learning.
\newline

\textbf{Question:}
\newline
What role does [module highlighted by blue bounding box] play in [module highlighted by red bounding box] estimation for online models?
\newline

\textbf{Ground-truth Answer:} 
\newline
Energy scores are used to classify unlabeled data into known and unknown categories in the first stage, and then further split unknown data into seen and unseen categories in the second stage. This is done by calculating the energy scores using a Gaussian Mixture Model to identify which cluster a sample belongs to, ultimately facilitating the estimation of seen and unseen data in the online model.
\newline

\small \textbf{Relevant Paper Context:} 
\newline
Section 3.3\hspace{3mm}Energy-Guided Discovery to Identify Unknowns:
\newline
\emph{``In Figure 3, we found that the two-stage split of the CGCD struggled to identify both known and unknown categories in our proposed online learning scenario. Online learning with batch-wise data led to severe forgetting in the split network, resulting in poor detection of unknown samples. In contrast, energy-based discovery showed better performance in novel category discovery without extra parameters compared to CGCD. Moreover, as it does not require an additional learning phase, energy-based discovery enables end-to-end training. Inspired by this observation, we propose an energy-guided discovery approach for novel category discovery. To the best of our knowledge, this is the first work to utilize the energy score for novel category discovery. 
\newline
The process in the second stage is the same as the first stage, where we split the unknown data into seen and unseen categories based on their energy scores. For the initial batch of incremental sessions, the online model is identical to the offline model. Additionally, since the initial batch data is the first data of the incremental session, any data classified as unknown is assumed to be unseen data. The proposed energy-guided discovery splits unlabeled data into known, seen, and unseen categories effectively. Unlike prior methods which require sufficient data, it can identify novel categories with batch-wise data.''}

\end{blueBox}
\clearpage

%% file: appendix/scenario_examples/Literature_Background.tex
\subsection{Literature Background Diagram example}\label{app:e3}
\begin{blueBox}[Literature Background]
\begin{center}
    \includegraphics[width=0.8\linewidth]{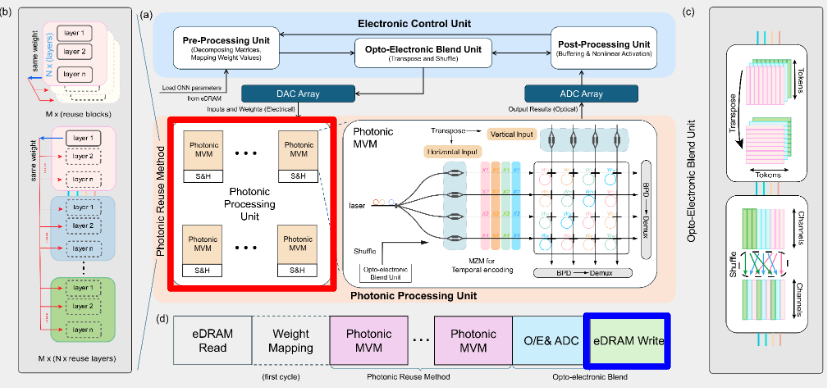}
\end{center}

\small\textbf{Paper Title:}
\newline
Reuse and Blend: Energy-Efficient Optical Neural Network Enabled by Weight Sharing
\newline

\textbf{Figure Caption:}
\newline
(a) Overview of the R\&B architecture. Each PPU contains a photonic MVM unit and a sampling and hold (S\&H) unit. (b) Photonic Reuse Method (PRM). Block-wise reuse allows weight sharing among blocks (a block typically contains multiple layers). Layer-wise reuse enables weight sharing between individual layers. (c) Opto-electronic Blend Unit (OBU). OBUs handle shuffle operations via the peripheral circuit and perform transpose operations directly in the optical domain. (d) Computing pipeline of our R\&B architecture.
\newline

\textbf{Question:}
\newline
What is the relationship between [module highlighted by red bounding box] and [module highlighted by blue bounding box]?
\newline

\textbf{Ground-truth Answer:} 
\newline
The Photonic Processing Unit (PPU) plays a role in the R\&B architecture's computation process where inputs are initially retrieved from eDRAMs to be processed by PPUs. After processing, the outputs are stored back into eDRAMs for the next layer computation.
\newline

\small \textbf{Relevant Paper Context:} 
\newline
Section 3.3\hspace{3mm}Computing Pipeline: 
\newline
\emph{``The computation process of our R\&B architecture contains three stages, as illustrated in Fig. 2(a). Initially, inputs are retrieved from eDRAMs, and corresponding weights are allocated to MRRs. Subsequently, following the PRM configuration, the weights are fixed and reused, allowing the inputs to pass through the MRRs to be optically weighted. The intermediate MVM results generated by the PPUs are then detected by BPDs, where they are converted into summed currents and digitized by ADCs. In the final stage, OBUs transform these outputs to generate the layer-wise results, which are then stored back in eDRAMs in preparation for the next computational layer.\newline
A critical aspect of this architecture is the role of the OBU during inference, mirroring its function during training and inference by executing essential shuffle and transpose operations. Subsequently, following the PRM configuration, the weights are fixed and reused. Along with PRM, these two technologies constitute the primary innovation of our R\&B architecture. By leveraging one MRR array to represent multiple weight matrices, the architecture dramatically reduces the frequency of MRR writing operations, along with power consumption and latency, all while sustaining high performance.''}

\end{blueBox}
\clearpage

%% file: appendix/scenario_examples/Literature_Background2.tex
\begin{blueBox}[Literature Background]
\begin{center}
    \includegraphics[width=0.8\linewidth]{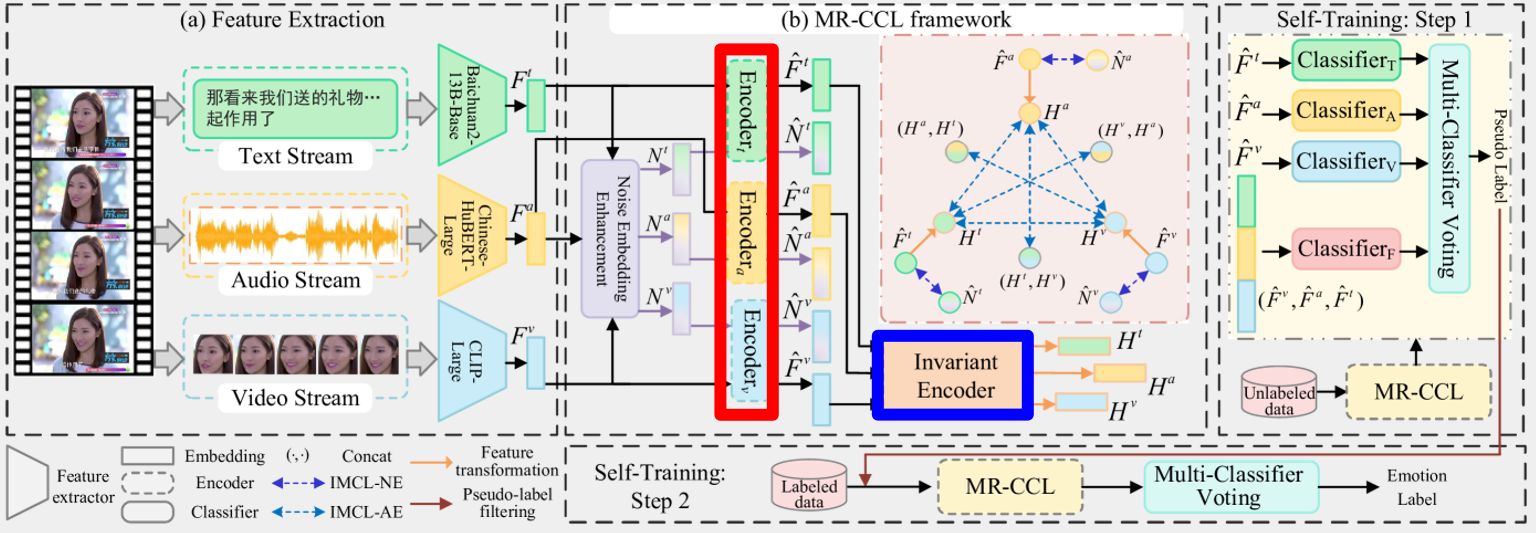}
\end{center}

\small\textbf{Paper Title:}
\newline
Leveraging Contrastive Learning and Self-Training for Multimodal Emotion Recognition with Limited Labeled Samples
\newline

\textbf{Figure Caption:}
\newline
Illustration of our MR-CCL framework.

\textbf{Question:}
\newline
How does the interaction between [module highlighted by red bounding box] and [module highlighted by blue bounding box] support contrastive learning methodology?
\newline

\textbf{Ground-truth Answer:} 
\newline
The interaction between invariant encoder and modality-specific encoders supports contrastive learning by enabling the extraction of both modality-specific embeddings and modality-invariant embeddings. Modality-specific encoders, built with Transformer layers, encode characteristics unique to each modality, while the invariant encoder, structured with linear layers, transforms these embeddings into a unified vector space, capturing relationships between modalities. This facilitates effective feature fusion and interaction for contrastive learning.
\newline

\small \textbf{Relevant Paper Context:} 
\newline
Section 2.4.1\hspace{3mm}Modality Representation Combinatorial Contrastive Learning: 
\newline
\emph{``We propose Modality Representation Combinatorial Contrastive Learning (MR-CCL) framework to leverage the given unlabeled data more effectively. The overall architecture is illustrated in Figure 2.
\newline
In the realm of multimodal emotion recognition tasks, a proven strategy involves extracting features specific to each modality as well as invariant features that capture relationships between different modalities, which facilitates the fusion and interaction of diverse modal inputs (Zuo et al. 2023; Liu et al. 2024b). To implement our framework, we begin by pre-training three specificity encoders and one invariant encoder using all available unlabeled data. This process is designed to enable the model to discern and encode the intrinsic characteristics and structural nuances of multimodal data. Each specificity encoder is composed of multiple Transformer layers, while the invariant encoder is structured with linear layers.''}

\end{blueBox}
\clearpage

%% file: appendix/scenario_examples/Experimental_Results.tex
\subsection{Experimental Results Diagram example}\label{app:e4}
\begin{blueBox}[Experimental Results]
\begin{center}
    \includegraphics[width=0.8\linewidth]{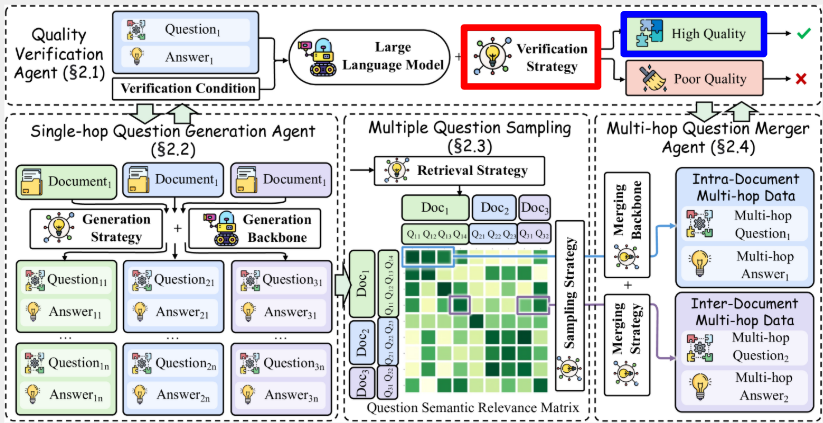}
\end{center}

\small\textbf{Paper Title:}
\newline
What are the Essential Factors in Crafting Effective Long Context Multi-Hop Instruction Datasets? Insights and Best Practices
\newline

\textbf{Figure Caption:}
\newline
The overall process of our Multi-agent Interactive Multi-hop Generation (MIMG) data synthesis framework.
\newline

\textbf{Question:}
\newline
How does the [module highlighted by red bounding box] ensure [module highlighted by blue bounding box]?
\newline

\textbf{Ground-truth Answer:} 
\newline
The Verification Strategy ensures high-quality data output by employing scoring and classification strategies. Scoring involves prompting LLMs to generate continuous scores and setting a threshold to determine high-quality data. This strategy captures the nuances of human judgments better than binary classification, which simply classifies samples as high-quality or not. This process ensures consistency, precision, and alignment with human judgment, thus improving data quality.
\newline

\small \textbf{Relevant Paper Context:} 
\newline
Section 2.1\hspace{3mm} Quality Verification Agent:
\newline
\emph{``The first module in our framework is Quality Verification Agent, which ensures that the generated questions and answers meet a certain standard of quality. This component involves two main processes:
\newline Verification Strategy:This includes additional heuristic strategies to judge which samples should be contained as high-quality data. This includes additional heuristic strategies to judge which samples should be contained as high-quality data. Specifically, we utilize two wide-used verification strategies:\newline • Scoring: We prompt LLMs to generate continuous scores, manually set a more reliable threshold score based on the validation set, and set those exceeding the threshold score as high-quality data.\newline • Classification: We prompt LLMs to generate binary classification and select those classified as high-quality data.
Verification Condition:
\newline Verification Condition:This involves setting specific conditions that both questions and answers must meet to be considered high-quality verification.The process includes:\newline • Criteria Perspectives: Criteria include relevance to the document, clarity, factual accuracy, logical coherence, and complexity of the question and answer.\newline • Auxiliary Context Information: We integrate additional contextual instructions to enhance the model’s accuracy and robustness, like guidelines. \newline • Auxiliary Generation Information: We enable the model to provide more reasoning rationale during output generation and observe whether this improves the robustness and accuracy of the verification process.''}
\newline

Section 3.1.1\hspace{3mm}Verification Strategy:
\newline
\emph{``Scoring is a Better Verification Strategy Compared with Classification.
As shown in Figure 3 (a), the scoring strategy shows significantly higher kappa and precision scores compared to binary quality classification. This statistical improvement suggests that scoring better captures the nuances of human judgments. This observation aligns with findings in short-context scenarios (Fu et al., 2024a), reinforcing the generalizability of scoring strategies across different lengths of textual data.''}

\end{blueBox}
\clearpage

%% file: appendix/scenario_examples/Experimental_Results2.tex
\begin{blueBox}[Experimental Results]
\begin{center}
    \includegraphics[width=0.8\linewidth]{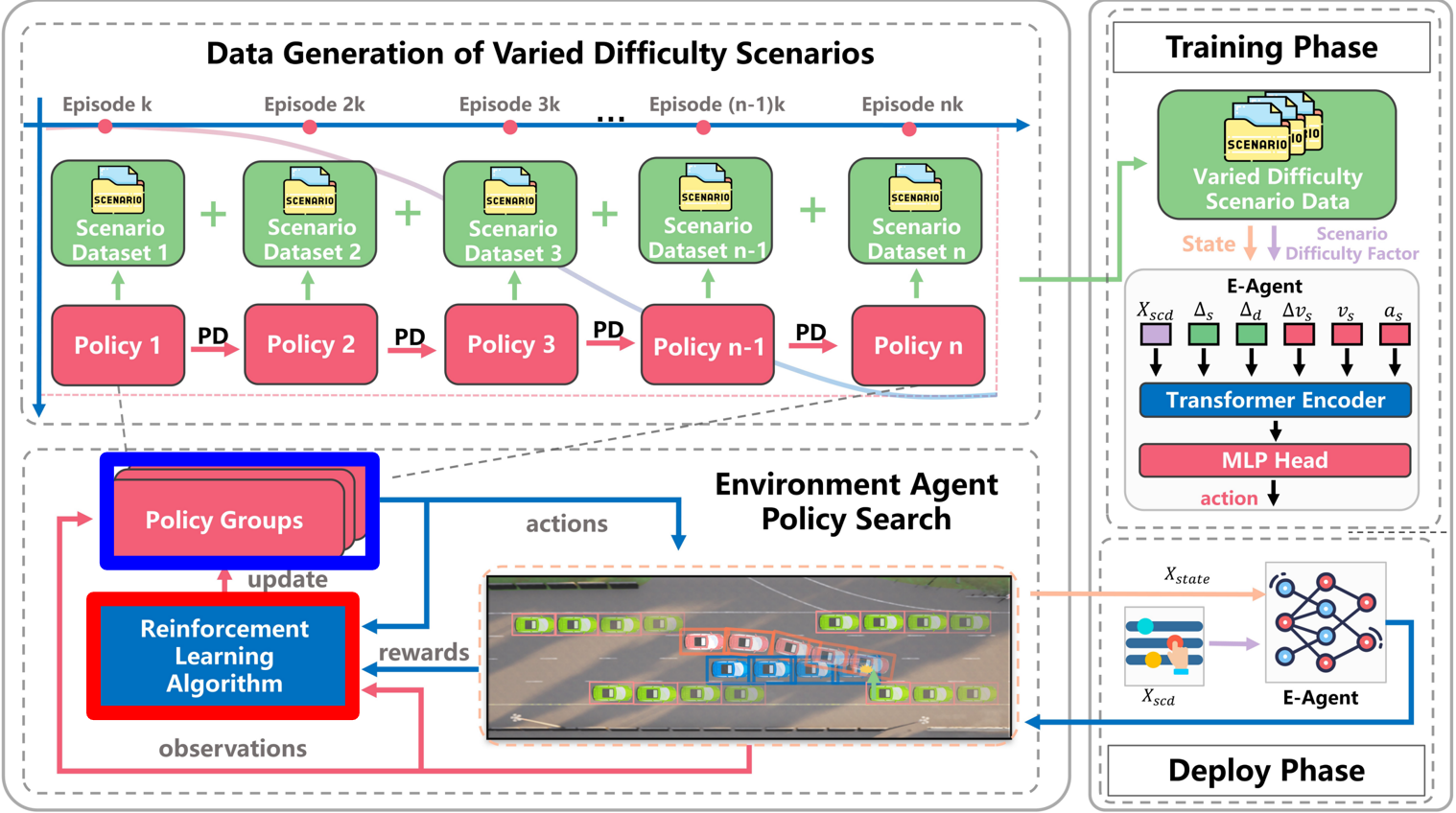}
\end{center}

\small\textbf{Paper Title:}
\newline
Quantitative Representation of Scenario Difficulty for Autonomous Driving Based on Adversarial Policy Search
\newline

\textbf{Figure Caption:}
\newline
Overall architecture of data driven quantitative representation method of scenario difficulty for autonomous driving based on environment agent policy search.
\newline

\textbf{Question:}
\newline
How does the [module highlighted by red bounding box]  contribute to the updates of  [module highlighted by blue bounding box]?
\newline

\textbf{Ground-truth Answer:} 
\newline
The reinforcement learning algorithm contributes to policy group updates by providing a systematic approach to search for optimal policies through the performance improvement phase, where model parameters are updated to pursue better performance, and the convergence stabilization phase, where the optimal policy is obtained. The performance at different stages is used to update and save model parameters to the constructed policy group.
\newline

\small \textbf{Relevant Paper Context:} 
\newline
Section 4\hspace{3mm} Data Generation of Scenarios with Varying Difficulty:
\newline
\emph{``In Section 3, we introduce the concept of environment agent to realize the adversarial policy search by combining logic rules with reinforcement learning. However, due to the black-box nature of data-driven methods, while adversarial actions can be generated, the difficulty of generating adversarial actions is difficult to quantify accurately, which limits the rationality of adversarial scenario generation.
\newline
In this section, a data generation method based on scenarios with varying difficulty is presented. The method uses the performance of different stages in the policy search convergence process as a reference to quantify the adversarial intensity, thereby achieving a quantitative representation of scenario difficulty. The model parameters of the environment agent trained on different stages are updated and saved, and then output to the constructed policy group. The policy group is used to generate data that forms the basis for training the scenario difficulty quantitative representation model..''}
\newline

Section 4.1\hspace{3mm}Verification Strategy:
\newline
\emph{``A reinforcement learning training process with stable convergence can be divided into two phases, i.e., the performance improvement phase and the convergence stabilization phase. In the performance improvement phase, the average return is still continuously increasing, which indicates that the policy search is still ongoing and the model parameters are still being updated to peruse better performance. In the convergence stabilization phase, however, the average return remains basically unchanged, indicating that the policy search is basically over, and the obtained policy is already the optimal policy that the current algorithm can achieve.
''}

\end{blueBox}
\clearpage

%% file: appendix/scenario_examples/Others.tex
\subsection{Other Diagram example}\label{app:e5}
\begin{blueBox}[Other Diagram Example]
\begin{center}
    \includegraphics[width=0.8\linewidth]{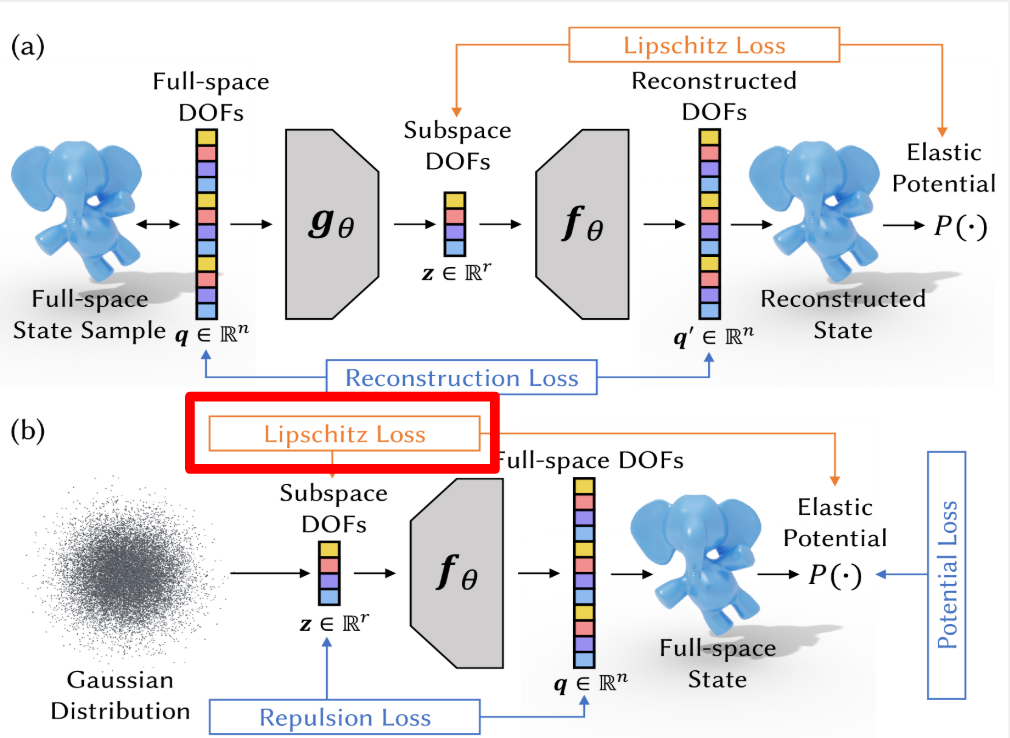}
\end{center}

\small\textbf{Paper Title:}
\newline
Accelerate Neural Subspace-Based Reduced-Order Solver of Deformable Simulation by Lipschitz Optimization
\newline

\textbf{Figure Caption:}
\newline
Network training settings for effective neural subspace construction. (a) The supervised setting. (b) The unsupervised setting. Conventional methods only consider the loss shown in blue but do not optimize the Lipschitz loss (shown in orange) to control the landscape of simulation objective in the subspace.
\newline

\textbf{Question:}
\newline
What are potential limitations of using  [module highlighted by red bounding box] in neural subspace training?
\newline

\textbf{Ground-truth Answer:} 
\newline
Potential limitations of using Lipschitz optimization in neural subspace training include the intractability of directly optimizing the Lipschitz constant due to the need to traverse all possible point pairs, sparse gradients that could damage Lipschitz characteristics in certain areas, increased memory usage, and potential memory shortages when training high-resolution meshes.
\newline

\small \textbf{Relevant Paper Context:} 

Section 5.4  \hspace{3mm} Limitation: 
\newline
\emph{``In this work, Lipschitz optimization is only applied to the elastic potential term of eq. 2. Since the nonlinear mapping is also involved in the inertia term, this may lower the convergence speed of the simulation involving dynamics. Considering that the inertia term is in quadratic form, the Hessian Lipschitz of the inertia term can be optimized by minimizing or bounding the Lipschitz constant of the network’s input-output Jacobian . This is a promising direction for future work to further accelerate the simulation with dynamics.
Another limitation of our method is the extended training time introduced by incorporating Lipschitz optimization into the pipeline. As shown in Table 1, even with cubature acceleration, the training time is still increased by approximately five times compared to the conventional method. This issue can be addressed by employing fast approximate methods to estimate Lipschitz energy.''}

\end{blueBox}
\clearpage

%% file: appendix/scenario_examples/Others2.tex
\begin{blueBox}[Other Diagram Example]
\begin{center}
    \includegraphics[width=0.8\linewidth]{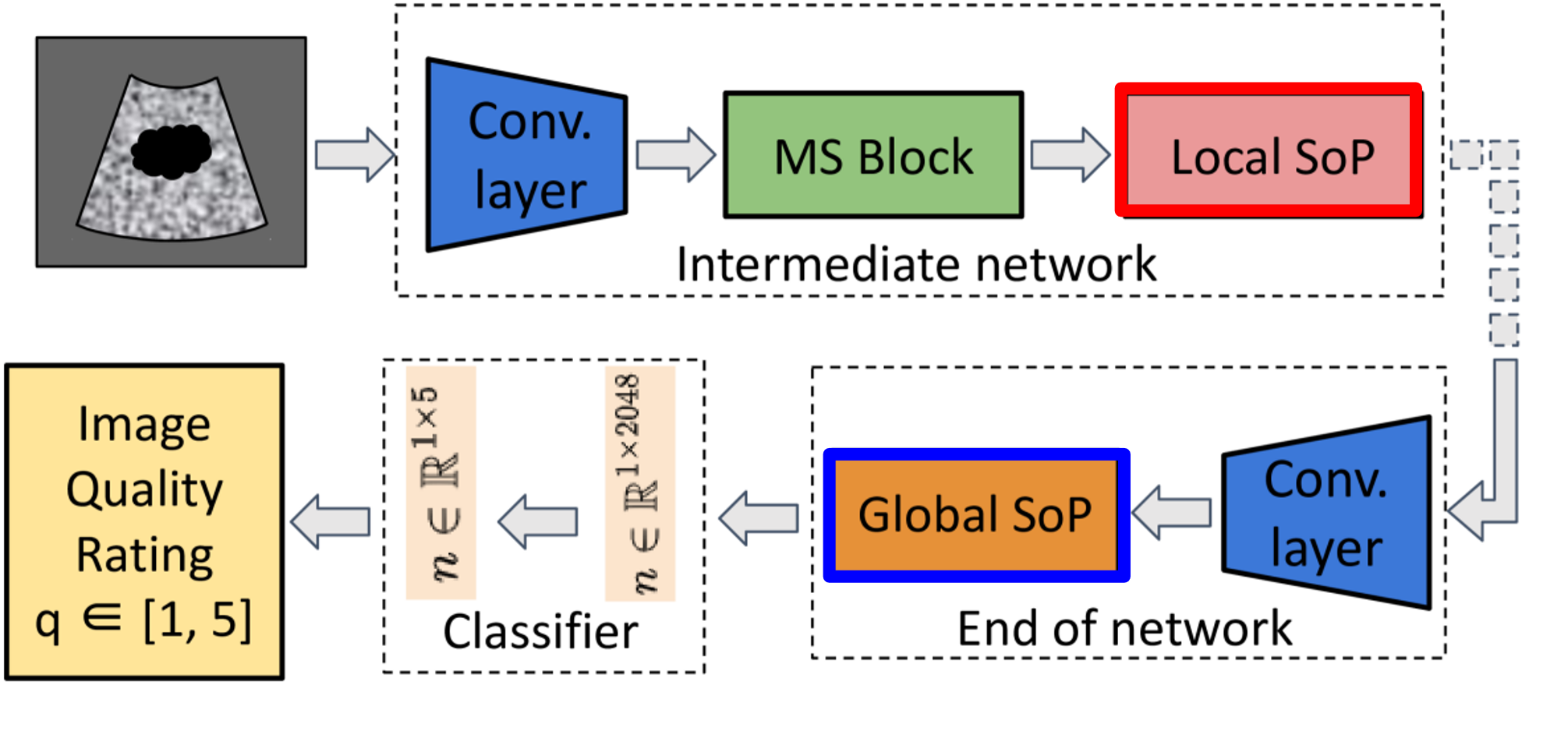}
\end{center}

\small\textbf{Paper Title:}
\newline
Coaching a Robotic Sonographer: Learning Robotic Ultrasound with Sparse Expert’s Feedback
\newline

\textbf{Figure Caption:}
\newline
State space representation using a deep convolution neural network
\newline

\textbf{Question:}
\newline
What are the potential challenges of combining [module highlighted by red bounding box] and [module highlighted by blue bounding box] in extracting meaningful image features?
\newline

\textbf{Ground-truth Answer:} 
\newline
Combining local and global second-order pooling (SoP) poses challenges such as increased computational complexity, potential feature redundancy, and the need for careful hyperparameter tuning. It demands substantial data to effectively handle multi-scale features while ensuring the model’s robustness. Additionally, balancing local and global information without conflicts can complicate optimization, particularly in real-time medical applications.
\newline

\small \textbf{Relevant Paper Context:} 

Section 2.1  \hspace{3mm} State space: 
\newline
\emph{``The state s is defined based on the ultrasound image. We have adopted an image quality classification network from our previous work , which used ResNet50 as a base network with multi-scale and higher-order processing of the image for conducting the holistic assessment of the image quality. The block diagram of this network is shown in Fig. 2. This classifier first extracts features at multiple scales to encode the inter-patient anatomical variations. Then, it uses second-order pooling (SoP) in the intermediate layers (local) and at the end of the network (global) to exploit the second-order statistical dependency of features. The local-to-global SoP will capture the higher-order relationships between different spatial locations and provide the seed for correlating local patches. This network encodes the image into a feature vector of size 
2048
, which represents the state of the policy.''}

\end{blueBox}
\clearpage

%% file: appendix/model_configuration.tex
\begin{table*}[h]
\centering
\footnotesize
\resizebox{\textwidth}{!}{%
\begin{tabular}{llllcrrc}
\toprule
\textbf{Organization} & \textbf{Model} & \textbf{Release} & \textbf{Version} & \textbf{\begin{tabular}[c]{@{}c@{}}Context\\Window\end{tabular}} &  \\
\midrule
\multicolumn{8}{c}{\emph{\textbf{Proprietary 
 Models}}} \\
 \midrule
\multirow{2}{*}{OpenAI} &o4-mini &2025-4 & \texttt{o4-mini-2025-04-16} & -- \\
& GPT-4.1 &2025-4 & \texttt{gpt-4.1-2025-04-14} & --\\
& GPT-4.1-mini &2025-4 & \texttt{gpt-4.1-mini-2025-04-14} & --\\
& GPT-4o &  2024-8  &  \texttt{gpt-4o-2024-08-06}  &-- \\

\noalign{\vskip 0.5ex}\hdashline\noalign{\vskip 0.5ex}

\multirow{1}{*}{Google} & Gemini-2.5-Flash & 2024-5 & \texttt{gemini-2.5-flash-preview-05-20} &--  \\
\noalign{\vskip 0.5ex}\hdashline\noalign{\vskip 0.5ex}

\midrule
\multicolumn{8}{c}{\emph{\textbf{Open-source Multimodal Foundation Models}}} \\
 \midrule

\multirow{2}{*}{Mistral AI} 
& Pixtral-12B & 2024-9 & \texttt{Pixtral-12B-2409} &128k \\
& Mistral-Small-3.1 & 2025-3 & \texttt{Mistral-Small-3.1-24B} &128k \\

\noalign{\vskip 0.5ex}\hdashline\noalign{\vskip 0.5ex}

\multirow{2}{*}{Microsoft} 
& Phi-3.5-Vision & 2024-7 & \texttt{Phi-3.5-vision-instruct} &32k\\
&Phi-4-Multimodal&2025-3&\texttt{Phi-4-Multimodal}&128k\\

\noalign{\vskip 0.5ex}\hdashline\noalign{\vskip 0.5ex}

\multirow{5}{*}{Shanghai AI Lab} 
& InternVL3-38B & 2025-04 & \texttt{InternVL-3-38B} & 32k\\
& InternVL3-8B & 2025-04 & \texttt{InternVL3-8B} & 32k\\
& InternVL2.5-38B & 2024-11 & \texttt{InternVL2.5-38B} & 32k\\
& InternVL2.5-8B & 2024-11 & \texttt{InternVL2.5-8B} & 32k\\
& InternVL2-8B & 2024-06 & \texttt{InternVL2-8B} & 32k\\

\noalign{\vskip 0.5ex}\hdashline\noalign{\vskip 0.5ex}

\multirow{4}{*}{Alibaba} 
& Qwen2.5-VL-72B & 2025-1 & \texttt{Qwen2.5-VL-72B-Instruct} &128k\\ 

& Qwen2-VL-72B & 2024-9 & \texttt{Qwen2-VL-72B-Instruct} &32k\\ 
& Qwen2.5-VL-7B & 2025-1 & \texttt{Qwen2.5-VL-7B-Instruct} &128k \\
& Qwen2-VL-7B & 2024-9 & \texttt{Qwen2-VL-7B-Instruct} &40k \\

\bottomrule
\end{tabular}
}
\caption{
Details of the multimodal foundation models evaluated in \ours. The ``Source'' column includes URLs for proprietary models and Hugging Face (HF) model names for open-source models.
}
\label{tab:model-info}
\end{table*}

%% file: figure_tex/cot_prompt.tex
\begin{figure}[H]
\begin{tcolorbox}[colback=black!3!white, colframe=black!50!white, title=Chain-of-Thought Prompt for Question Answering, fontupper=\footnotesize, fonttitle=\footnotesize]

\textbf{[System Input]} \\
\newline
You are a computer science researcher. You are provided a schematic diagram that provides an overview of the given paper's main research. Your task is to interpret the diagram and locate question-relevant information within the paper to answer the provided question. If the question cannot be answered based on the paper context, clearly conclude your response with: ``I do not know.''
\newline
\newline

\textbf{[User Input]}\\
\newline
\textcolor{gray}{\{image input: schematic\_diagram\}}\\
\newline
Schematic diagram's caption: \\
\textcolor{gray}{\{text input: caption\}} \\

Paper Context:\\
\textcolor{gray}{\{text input: paper\}}
\newline

Question: 
\textcolor{gray}{\{text input: question\}}
\newline

Follow the instructions and think step by step to answer the given question.
\end{tcolorbox}
\caption{The CoT prompt used in our experiment.
}
\label{fig:cot_prompt}
\end{figure}





%% file: appendix/error_examples/err1.tex
\subsection{Failure to interpret and contextualize schematic diagrams}\label{app:err1}
\begin{blueBox}[Literature Background]
\begin{center}
    \includegraphics[width=0.7\linewidth]{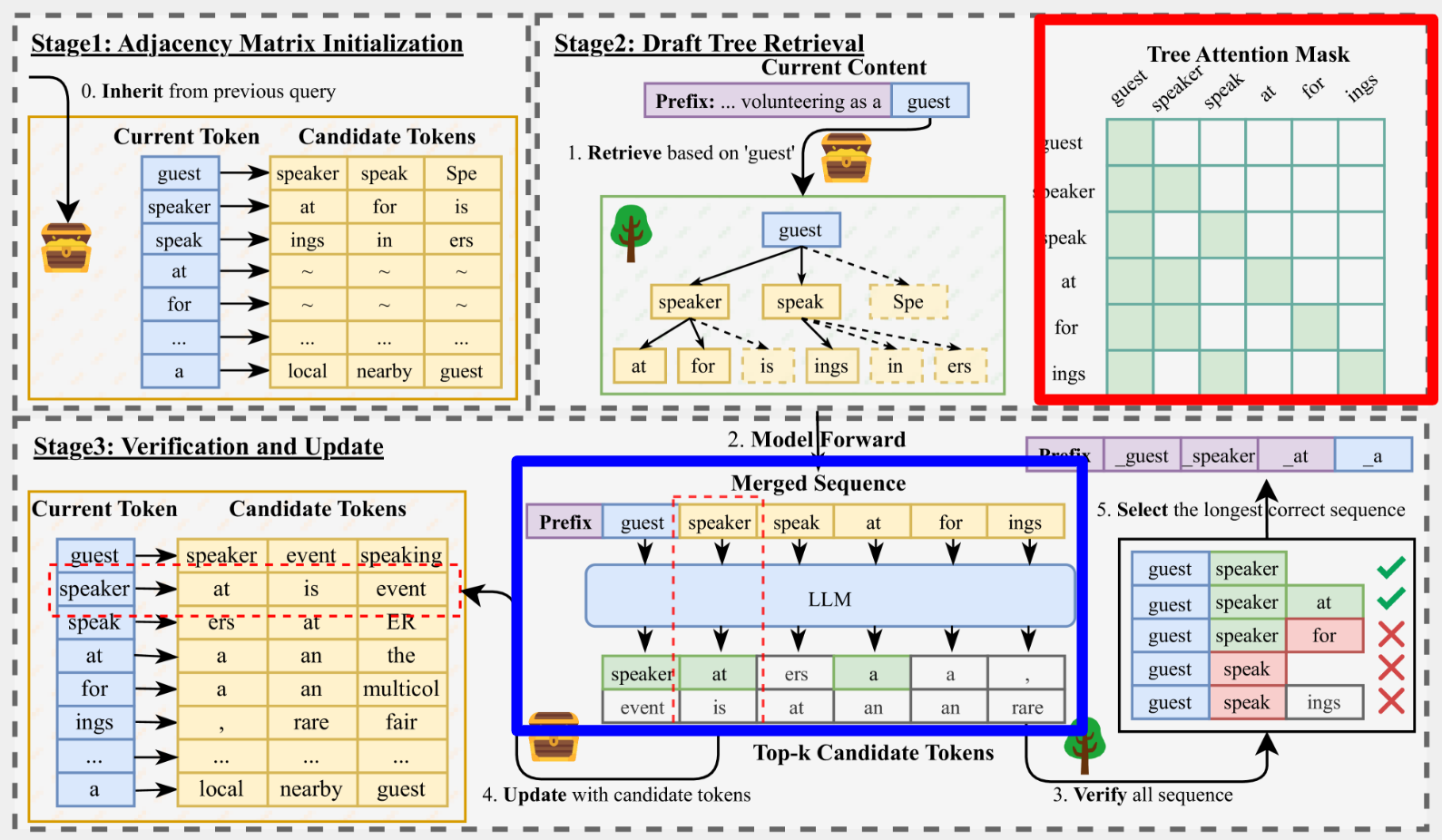}
\end{center}

\small \textbf{Paper Title:}
\newline
Turning Trash into Treasure: Accelerating Inference of Large Language Models with Token Recycling

\textbf{Figure Caption:}
\newline
An overview of Token Recycling. The adjacency matrix, initialized by inheriting from the previous query, stores candidate tokens. Token Recycling first retrieves a draft tree from the matrix based on the last token of the current content. The tree is then compressed into a merged sequence with a corresponding tree attention mask and sent to the LLM for a forward pass. After processing, all possible draft sequences are extracted and verified. The longest correct sequence is selected and added to the content. Finally, the top-k candidate tokens are used to update the matrix for the next iteration.

\textbf{Question:}
\newline
How does [module highlighted by red bounding box] interact with [module highlighted by blue bounding box] ?

\textbf{Ground-truth Answer:} 
\newline
The tree attention mask compresses multiple sequences into a single merged sequence while preserving a tree structure. Within this structure, each child node attends only to its parent nodes, preventing sibling tokens from interfering with each other. This allows the LLM to process and verify all possible sequences in parallel.

\small \textbf{Relevant Paper Context:} 
\newline
Section 2.3 \hspace{3mm} Tree Attention:
\newline
\emph{``The traditional causal attention masks are designed for linear sequences, where each token attends to all previous tokens, restricting speculative decoding to verifying one sequence at a time. However, as the sequence lengthens during draft token generation, the number of potential continuations increases. For example, in the draft tree in Figure 2, the token following 'guest' could be 'speaker' or 'speak', while both 'at' and 'for' could follow 'speaker'. This creates a need to verify multiple draft sequences simultaneously. Tree attention modifies the attention mask to address this by compressing multiple sequences into a single merged sequence, such as ['guest', 'speaker', 'speak', 'at', 'for', 'ings'], while preserving a tree structure in the tree attention mask. Each child node attends only to its parent nodes, preventing sibling tokens from interfering with each other. After the LLM processes the merged sequence, all possible sequences such as 'guest speaker', 'guest speaker at', 'guest speaker for', and 'guest speak', along with their corresponding output tokens, are extracted based on the tree structure and verified in parallel.''}

\end{blueBox}

\begin{wronganswer}
\small \textbf{Qwen2-VL-72B:}\\
The module highlighted by the red bounding box, labeled "Tree Attention Mask," interacts with the module highlighted by the blue bounding box, labeled "Merged Sequence," by providing attention weights to the tokens in the merged sequence. The tree attention mask determines which tokens in the merged sequence should be attended to during the model's forward pass. This allows the model to focus on relevant tokens and ignore irrelevant ones, improving the accuracy of the generated sequence.
\end{wronganswer}

\begin{correctanswer}
\small \textbf{Error Analysis:}\\
\small Qwen2-vl-72B's response fails primarily because it does not accurately capture the hierarchical relationship between the modules. It simplifies the interaction, describing the "Tree Attention Mask" (red box) as merely providing attention weights to the "Merged Sequence" (blue box), without explaining how the tree structure constrains and guides the sequence merging process. The response lacks clarity on how the red module enforces parent-child dependencies and prevents sibling token interference, which is crucial for maintaining the tree structure. Additionally, it does not highlight the step-by-step interaction or the layered collaboration between the two modules, leaving the explanation overly vague and incomplete.
\end{correctanswer}

\clearpage

\begin{blueBox}[Design Rationale]
\begin{center}
    \includegraphics[width=0.8\linewidth]{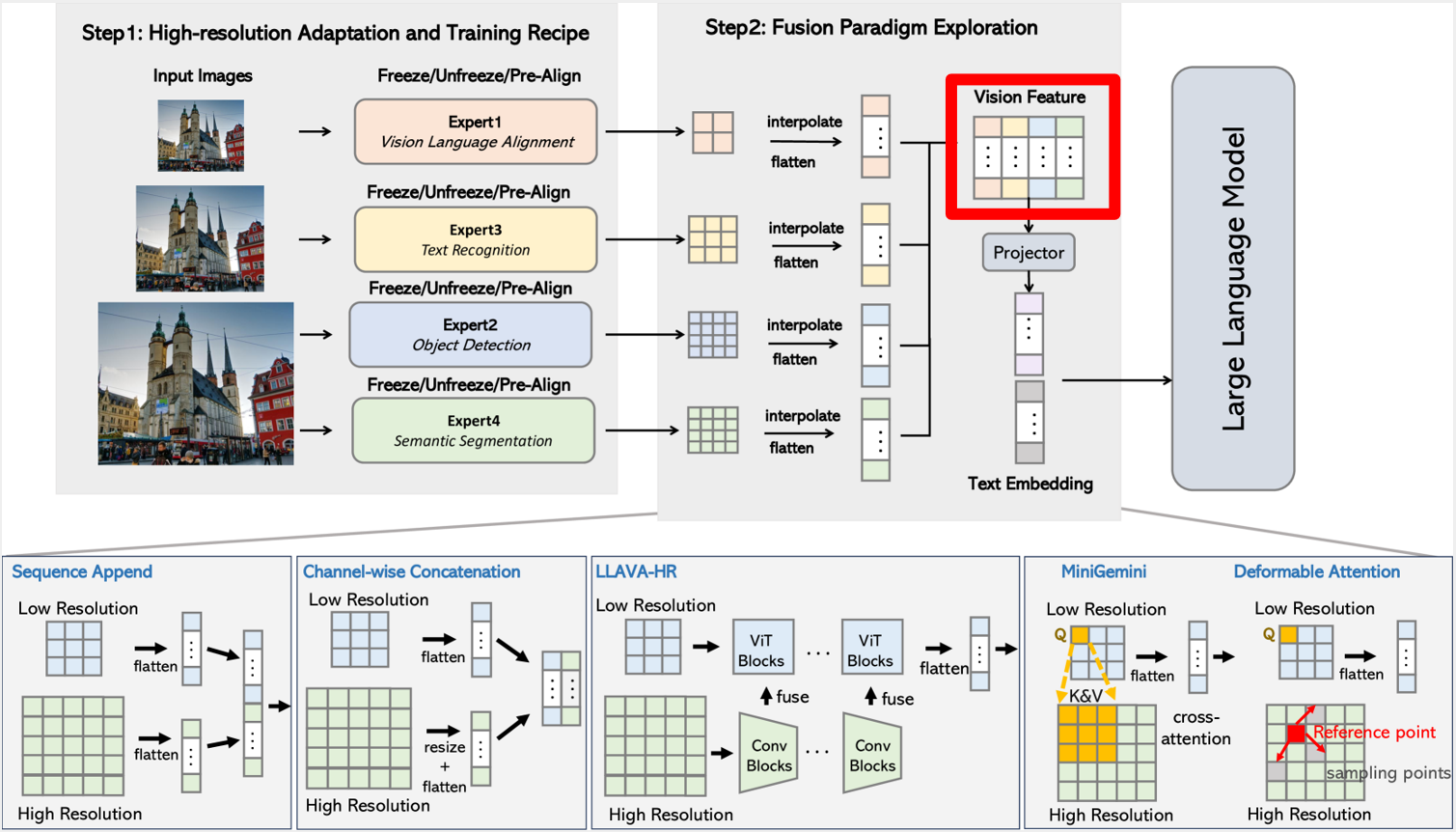}
\end{center}

\small \textbf{Paper Title:}
\newline
Eagle: Exploring The Design Space for Multimodal LLMs with Mixture of Encoders

\textbf{Figure Caption:}
\newline
Overview of the Eagle exploration pipeline. 

\textbf{Question:}
\newline
What are the benefits of using channel-wise concatenation in the processing of [module highlighted by red bounding box]?

\textbf{Ground-truth Answer:} 
\newline
The benefits of using channel-wise concatenation in vision feature processing include achieving the best average performance, maintaining better throughput compared to sequence append, and offering performance, expandability, and efficiency.

\small \textbf{Relevant Paper Context:} 
\newline
Section 2.4 \hspace{3mm} Fusion strategy:
\newline
\emph{``We notice that existing popular fusion strategies, despite their variations in designs, can be broadly represented by the following several categories: (1) Sequence Append: directly appending the visual tokens from different backbones as a longer sequence; (2) Channel Concatenation: concatenating the visual tokens along the channel dimension without increasing the sequence length; (3) LLaVA-HR: injecting high-resolution features into low-resolution vision encoders using mixture-of-resolution adapter; (4) Mini-Gemini: using the CLIP tokens as the low resolution queries to cross-attend another high-resolution vision encoder in the co-located local windows. 
\newline
Although sequence append shows comparable performance to channel concatenation, it faces the challenge to handle more vision encoders due to the increasing sequence length. Hence, we choose direct channel concatenation as our fusion strategy considering its performance, expandability, and efficiency.''}

\end{blueBox}

\begin{wronganswer}
\small \textbf{Qwen2-VL-72B:}\\
Channel-wise concatenation allows for the combination of features from different resolutions, which can improve the performance of the model.
\end{wronganswer}

\begin{correctanswer}
\small \textbf{Error Analysis:}\\
\small Qwen's response lacks depth as it fails to interpret the schematic diagram and contextualize channel-wise concatenation within the broader design framework. It does not compare its benefits with other fusion strategies like sequence append, missing key insights on performance, expandability, and efficiency. A better response would explicitly reference the diagram and relate the highlighted module to the findings in the paper.
\end{correctanswer}

%% file: appendix/error_examples/err2.tex
\subsection{Inability to retrieve relevant context}\label{app:err2}
\begin{blueBox}[Experimental Results]
\begin{center}
    \includegraphics[width=0.8\linewidth]{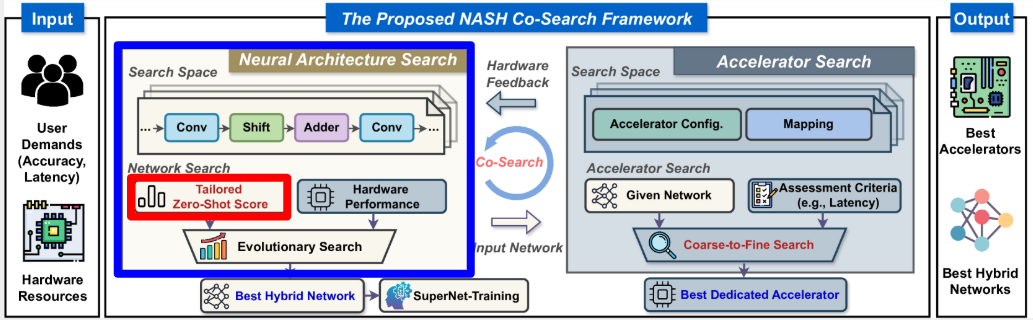}
\end{center}

\small\textbf{Paper Title:}
\newline
What are the Essential Factors in Crafting Effective Long Context Multi-Hop Instruction Datasets? Insights and Best Practices

\textbf{Figure Caption:}
\newline
The overview of our NASH framework, where we integrate both the neural architecture search (NAS) and coarse-to-fine accelerator search to directly obtain optimal pairing of models and accelerators. Specifically, the NAS consists of a tailored zero-shot metric to pre-identify promising multiplication-reduce hybrid models before supernet training. Besides, the accelerator search involves a novel coarse-to-fine search strategy to expedite the accelerator search process.

\textbf{Question:}
\newline
How does the [module highlighted by red bounding box] contribute to the efficiency of [module highlighted by blue bounding box]?

\textbf{Ground-truth Answer:} 
\newline
The tailored zero-shot score contributes to neural architecture search efficiency by enabling faster assessment due to its swift computational speed. The computation of the tailored zero-shot metric is significantly faster than assessing the accuracy of individual hybrid models derived from the supernet, leading to enhanced search efficiency.

\small \textbf{Relevant Paper Context:} 
\newline
Section 3.1 \hspace{3mm} The Tailored Zero-shot Metric:
\newline
\emph{``To enable a more accurate assessment of our hybrid networks, we integrate two selected zero-shot metrics. Given the significant difference in score magnitudes between these metrics, as shown in Figures 3(b) and 3(c), we focus on relative rankings rather than score magnitudes. Specifically, for a group of networks, the score of our tailored zero-shot metric for a specific network is determined by the relative ranking of its Zen-Score within the group. For instance, if a network exhibits the highest Zen-Score, its term yields a value of 1. The effectiveness of our tailored metric is validated through Table II and Figure 3, which demonstrate the highest Kendall-Tau Correlation. Additionally, this metric contributes to enhanced search efficiency due to the swift computational speed of both NN-Degree and Zen-Score. For example, assessing accuracy for an individual hybrid model from our supernet takes an average of several seconds, whereas computing our tailored zero-shot metric requires less time, making it over X times faster when tested on CIFAR100 and profiled on an NVIDIA GeForce RTX 2080Ti.''}
\newline

TABLE II \hspace{3mm} The Kendall Tau Coefficient between zero-shot metric scores and real model accuracy. T and E are the abbreviations for trainability and expressivity

| Metrics\hspace{22mm}             | Mult.-based | Mult.-reduced | Class |\newline
|--------------------------------|----------------|-------------------|--------|\newline
| SNIP\hspace{24.8mm}      | 0.40\hspace{9.5mm}        | -0.01\hspace{12.1mm}         | T\hspace{4.3mm}     |\newline
| Jacobian Covariant\hspace{7.5mm}     | 0.43\hspace{9.5mm}        | 0.06\hspace{13.1mm}          | T\hspace{4.3mm}     |\newline
| Grad Norm\hspace{17.3mm}              | 0.36\hspace{9.5mm}        | -0.09\hspace{12.1mm}         | T\hspace{4.3mm}      |\newline
| Synflow\hspace{21.5mm}                | 0.47\hspace{9.5mm}        | 0.06\hspace{13.1mm}          | T\hspace{4.3mm}      |\newline
| Grasp\hspace{24.3mm}                  | 0.35\hspace{9.5mm}        | 0.03\hspace{13.1mm}          | T\hspace{4.3mm}      |\newline
| Fisher\hspace{24mm}                 | 0.45\hspace{9.5mm}        | -0.05\hspace{12.1mm}         | T\hspace{4.3mm}      |\newline
| NN-Degree \hspace{16.5mm}             | 0.43\hspace{9.5mm}        | 0.32\hspace{13.1mm}          | T\hspace{4.3mm}      |\newline
| Linear Region Number\hspace{2.8mm}   | 0.32\hspace{9.5mm}        | 0.19\hspace{13.1mm}          | E\hspace{4.3mm}      |\newline
| Zen-Score \hspace{18.1mm}             | 0.40\hspace{9.5mm}        | 0.33\hspace{13.1mm}          | E\hspace{4.3mm}      |\newline
| Ours \hspace{25.1mm}                  | -\hspace{14mm}           | 0.47\hspace{13.1mm}          | T\&E\textbf{\hspace{1.6mm} }   |

\end{blueBox}

\begin{wronganswer}
\small\textbf{Qwen2-VL-72B:}\\
The Tailored Zero-Shot Score module contributes to the efficiency of the Neural Architecture Search module by providing a fast and accurate way to evaluate the performance of different network architectures without the need for full training. This allows the Evolutionary Search algorithm to efficiently explore the search space and find the best hybrid network.

\end{wronganswer}

\begin{correctanswer}
\small\textbf{Error Analysis:}\\
Qwen2's response fails to explicitly connect the Tailored Zero-Shot Score with the efficiency of Neural Architecture Search (NAS) in the manner suggested by the diagram and the surrounding context. Instead, it merely retrieves a generalized answer based on the keyword, without integrating the broader relationships or contextual dependencies illustrated in the diagram.

\end{correctanswer}

\clearpage
\begin{blueBox}[Literature Background]
\begin{center}
    \includegraphics[width=0.8\linewidth]{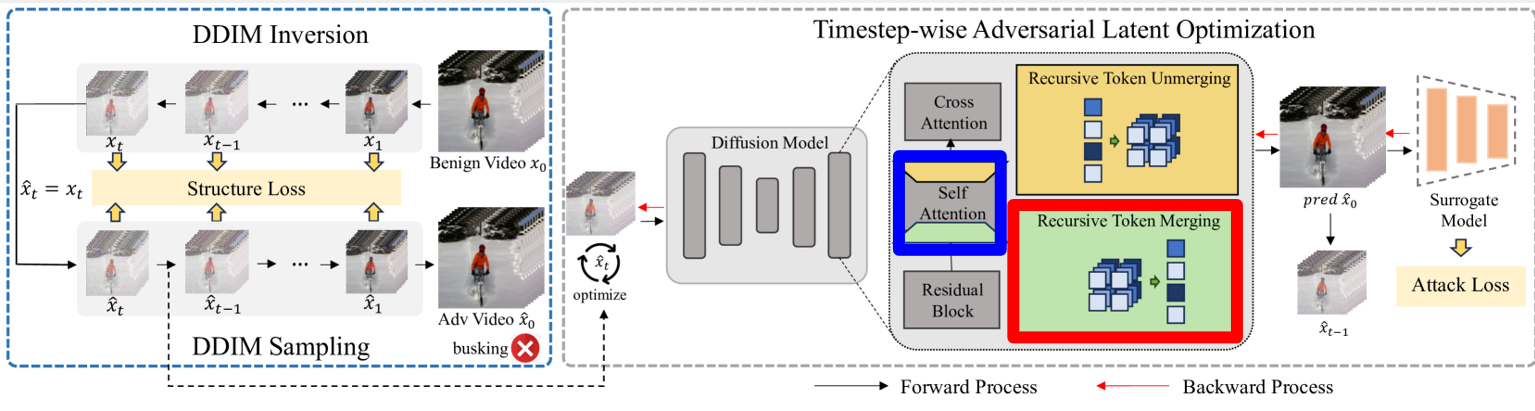}
\end{center}

\small\textbf{Paper Title:}
\newline
ReToMe-VA: Recursive Token Merging for Video Diffusion-based Unrestricted Adversarial Attack

\textbf{Figure Caption:}
\newline
Framework overview of the proposed ReToMe-VA. For a video clip, DDIM inversion is applied to map the benign frames into the latent space. Timestep-wise Adversarial Latent Optimization is employed during the DDIM sampling process to optimize the latents. Throughout the whole pipeline, Recursive Token Merging and Recursive Token Unmerging Modules are integrated into the diffusion model to enhance its effectiveness. Additionally, structure loss is utilized to maintain the structural consistency of video frames. Ultimately, the resulting adversarial video clip is capable of deceiving the target model.

\textbf{Question:}
\newline
How does [module highlighted by red bounding box] interact with [module highlighted by blue bounding box] to enhance video consistency?

\textbf{Ground-truth Answer:} 
\newline
Recursive Token Merging interacts with the Self Attention module by recursively matching and merging similar tokens across frames, enabling the Self Attention module to extract consistent features. This method ensures that the most similar tokens share identical outputs, which maximizes internal uniformity of features across frames and preserves temporal consistency, thereby enhancing video consistency.

\small \textbf{Relevant Paper Context:}
\newline
Section 3.3 \hspace{3mm} Recursive Token Merging:
\newline
\emph{TALO strategy perturbs each benign frame of video separately. This per‑frame optimization makes the frames likely optimized along different adversarial directions resulting in motion discontinuity and temporal inconsistency. Furthermore, separately perturbing each benign frame reduces the monotonous gradients because the interactions among the frames are not exploited. To this end, we introduce a recursive token merging (ReToMe) strategy that recursively matches and merges similar tokens across frames together enabling the self‑attention module to extract consistent features. In the following, we first provide the basic operation of token merging and token unmerging and then our recursive token merging algorithm.
\newline
Token Merging (ToMe) is first applied to speed up diffusion models through several diffusion‑specific improvements. Generally, tokens $T$ are partitioned into a source (\textit{src}) and destination (\textit{dst}) set. Then, tokens in \textit{src} are matched to their most similar token in \textit{dst}, and the $r$ most similar edges are selected subsequently. Next, we merge the connected $r$ most similar tokens in \textit{src} to \textit{dst} by replacing them with the linked \textit{dst} tokens. To keep the token number unchanged, we divide merged tokens after self‑attention by assigning their values to merged tokens in \textit{src}.
\newline
A self‑attention module takes a sequence of input and output tokens across all frames. To partition tokens across frames into \textit{src} and \textit{dst}, we define the stride as $B$. We randomly choose one out of the first $B$ frames (e.g., the $g$‑th frame) and select the subsequent frames every $B$ interval into the \textit{dst} set.
}

\end{blueBox}

\begin{wronganswer}
\small\textbf{Qwen2-VL-72B:}\\
The module highlighted by the red bounding box, which is the Recursive Token Merging block, interacts with the module highlighted by the blue bounding box, which is the Self Attention block, to enhance video consistency. The Recursive Token Merging block merges tokens from adjacent timesteps, while the Self Attention block attends to the merged tokens to ensure that the information from different timesteps is consistent. This interaction helps to maintain the temporal coherence of the video frames.

\end{wronganswer}

\begin{correctanswer}
\small\textbf{Error Analysis:}\\
Qwen2-VL-72B's response oversimplifies the interaction between Recursive Token Merging (ReToMe) and the Self Attention module. While it correctly identifies their roles, it fails to emphasize how ReToMe recursively merges similar tokens across frames, ensuring temporal consistency and internal feature uniformity. This process mitigates per-frame inconsistencies in adversarial training, preserving structural coherence. The response lacks depth in explaining how ReToMe enhances adversarial effectiveness and video consistency beyond simple token merging.

\end{correctanswer}

%% file: appendix/error_examples/err3.tex
\subsection{Reasoning error}\label{app:err3}

\begin{blueBox}[Literature Background]
\begin{center}
    \includegraphics[width=0.6\linewidth]{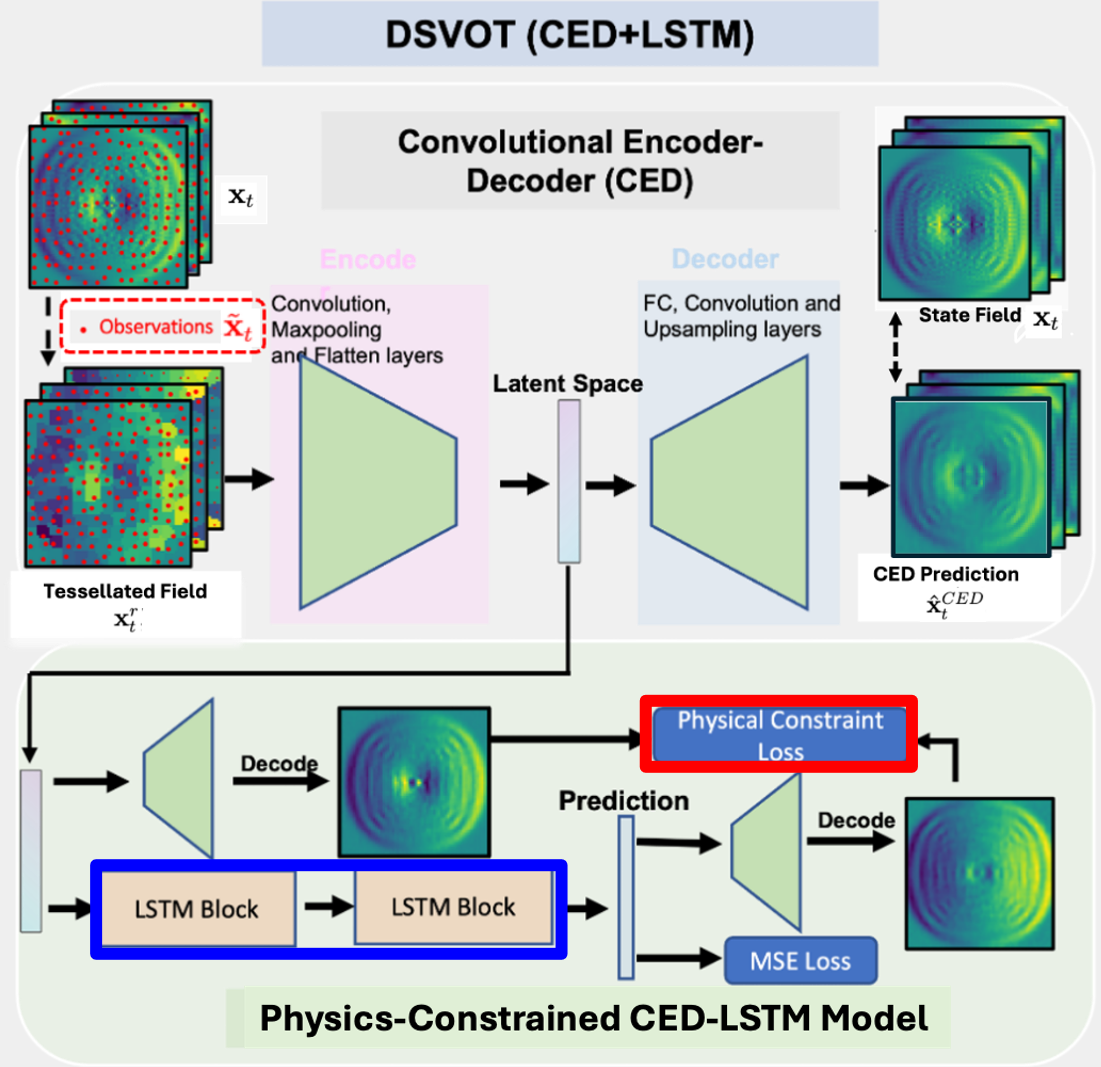}
\end{center}

\small\textbf{Paper Title:}
\newline
Dynamical system prediction from sparse observations using deep neural networks with Voronoi tessellation and physics constraint
\newline
\textbf{Figure Caption:}
\newline
Schematic representation of physics-constrained CED-LSTM model employing Voronoi tessellation for enhanced state field mapping from sparse observations.

\textbf{Question:}
\newline
What impact does incorporating  [module highlighted by red bounding box] have on the predictions of [module highlighted by blue bounding box]?

\textbf{Ground-truth Answer:} 
\newline
Incorporating physical constraint loss into LSTM block predictions enhances the accuracy and ensures physical consistency by making the model adhere to fundamental physical laws such as energy conservation, leading to more realistic prediction outcomes.

\small \textbf{Relevant Paper Context:} 
\newline
Section 2.5: \hspace{3mm}Physics Constraints:
\newline
\emph{``Energy conservation asserts that in a conservative system, the total energy remains constant over time. This concept is particularly relevant in systems where external energy exchanges are absent. To quantify alignment with energy conservation principles, we define an energy conservation loss function, energy, which measures the discrepancy between the energy states of the input and output fields. This function is integrated into the overall loss function to enhance the adherence of the model to energy conservation.''}

\end{blueBox}

\begin{wronganswer}
\small\textbf{Qwen2-VL-72B:}\\
Incorporating the physical constraint loss has a regularization effect on the predictions of the LSTM block. This is because the physical constraint loss ensures that the predictions made by the LSTM block are consistent with the underlying physics of the system being modeled.

\end{wronganswer}

\begin{correctanswer}
\small\textbf{Error Analysis:}\\
Qwen2-VL-72B's response lacks depth in its reasoning by overly simplifying the role of physical constraint loss as mere regularization. It fails to acknowledge the iterative parameter update mechanism and how it enhances adaptability in dynamic systems. Additionally, the explanation does not address the fundamental issue of data drift and evolving task complexities, which are critical in ensuring long-term model robustness and accuracy.

\end{correctanswer}
\clearpage

%% file: appendix/error_examples/err4.tex
\subsection{Overconfident response to unanswerable questions}\label{app:err4}
\begin{blueBox}[Implementation Details]
\begin{center}
    \includegraphics[width=0.8\linewidth]{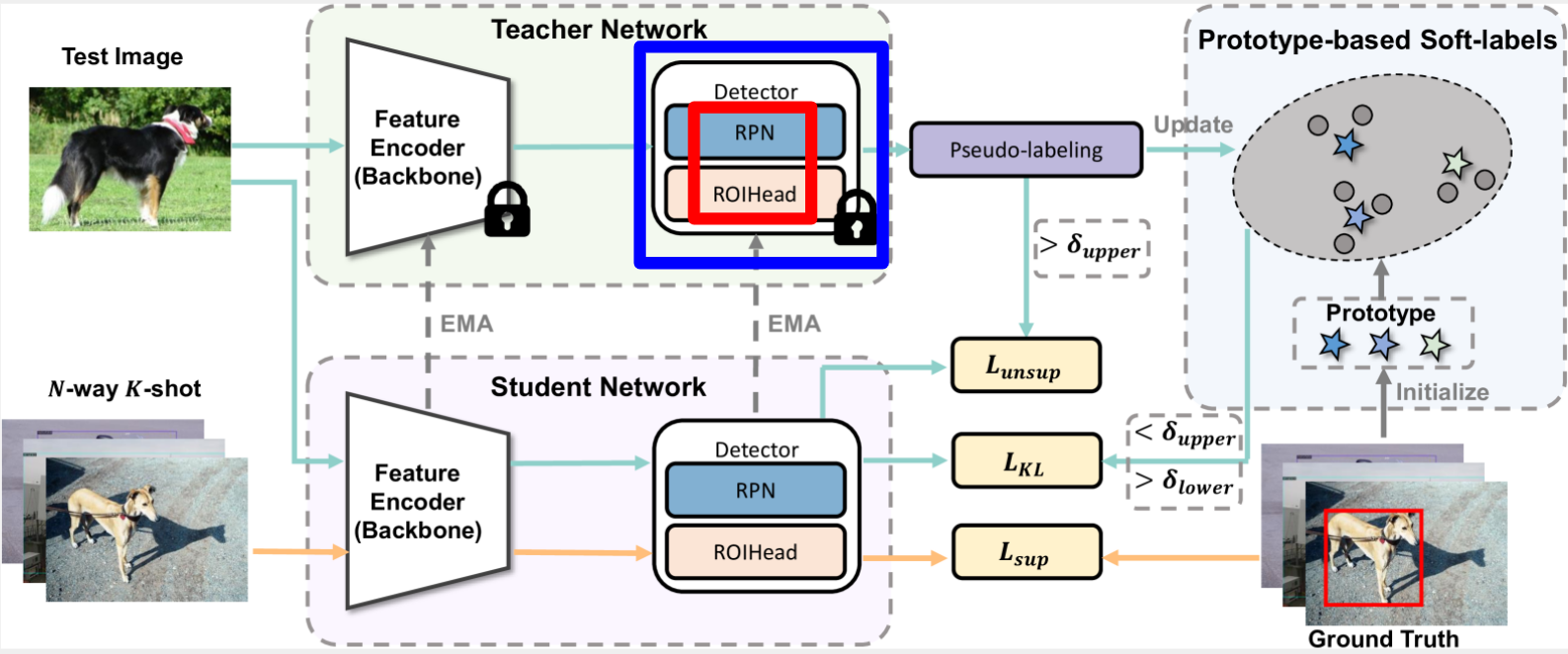}
\end{center}

\small\textbf{Paper Title:}
\newline
PS-TTL: Prototype-based Soft-labels and Test-Time Learning for Few-shot Object Detection
\newline
\textbf{Figure Caption:}
\newline
The overview of the proposed Prototype-based Soft-labels and Test-Time Learning (PS-TTL) framework for FSOD. Both the student and teacher networks are first initialized by the few-shot detector and then fine-tuned on test data. The teacher network takes test data as input to generate pseudo-labels, while the student model is trained using these pseudo-labels after post-processing with N-way K-shot data as supervision signals and updates the teacher net- work through EMA. A Prototype-based Soft-labels (PS) strategy is adopted to maintain class prototypes and compute the feature similarity between low-confidence pseudo-labels and class prototypes to replace them with soft-labels.

\textbf{Question:}
\newline
What are the specific functions of the [module highlighted by red bounding box] in the [module highlighted by blue bounding box]? 
\newline
\textbf{Ground-truth Answer:} 
\newline
The question is unanswerable.

\small \textbf{Relevant Paper Context:} 
\newline
Section 2.2 \hspace{3mm} Few-shot Object Detection:
\newline
\emph{``Similarly, Meta R-CNN combines a two-stage detector and reweights RoI features in the detection head. Attention-RPN exploits matching relationship between the few-shot support set and query set with a contrastive training scheme, which can then be applied to detect novel objects without retraining and fine-tuning.''}
\end{blueBox}

\begin{wronganswer}
\small\textbf{Qwen2-VL-72B:}\\
The module highlighted by the red bounding box includes the RPN and the ROIHead, which are components of the module highlighted by the blue bounding box, the Detector. The ROIHead is responsible for generating region proposals, while the RPN is responsible for classifying and regressing the proposals.

\end{wronganswer}

\begin{correctanswer}
\small\textbf{Error Analysis:}\\
Qwen2-vl-72B's response is problematic because the question is explicitly unanswerable, as confirmed by the official answer. The model incorrectly generates a detailed explanation by hallucinating the roles of modules (e.g., RPN and ROIHead) without any supporting information.

\end{correctanswer}
\newpage

\begin{blueBox}[Literature Background]
\begin{center}
    \includegraphics[width=0.8\linewidth]{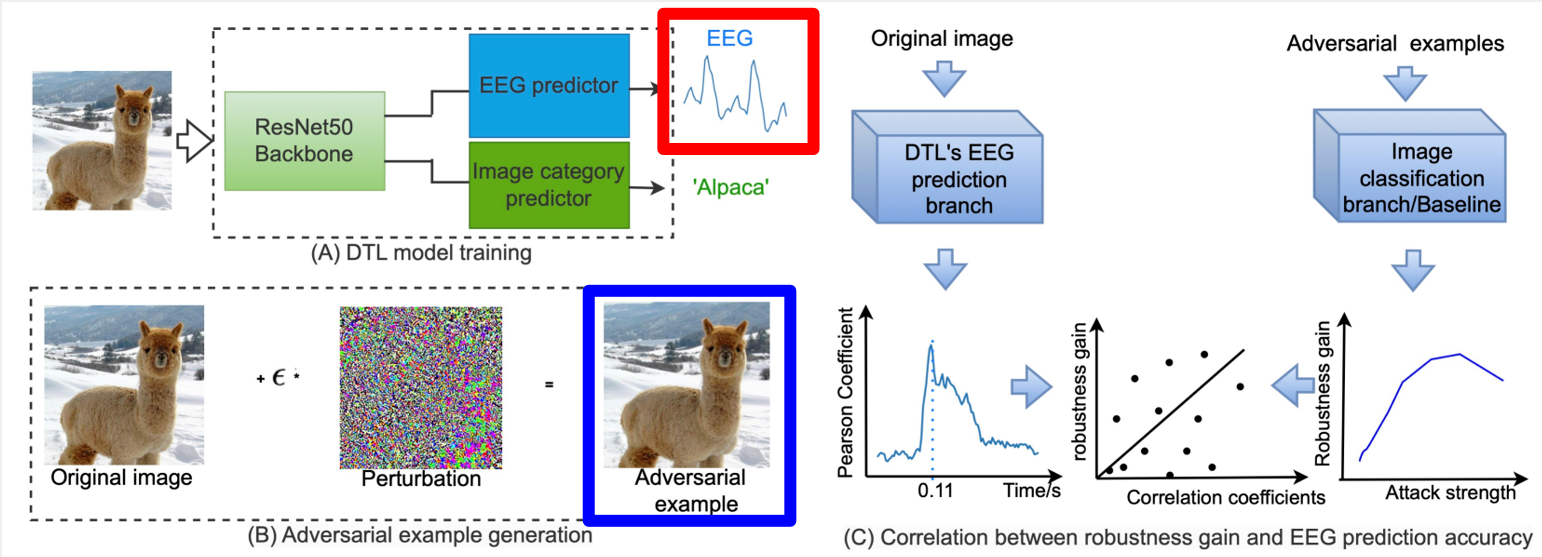}
\end{center}

\small\textbf{Paper Title:}
\newline
Limited but consistent gains in adversarial robustness by co-training object recognition models with human EEG
\newline
\textbf{Figure Caption:}
\newline
Paradigm for improving adversarial robustness via co-training with human EEG: We first trained dual-task learning (DTL) models with original and shuffled EEG data and then evaluated their robustness against various adversarial attacks. We trained four clusters of ResNet50 backbone models, each incorporating a different independent EEG predictor: Dense Layers (CNN), Recurrent Neural Networks (RNN), Transformer, and Attention layers. Finally, we measured the relationship between adversarial robustness gain and EEG prediction accuracy.

\textbf{Question:}
\newline
How are [module highlighted by red bounding box] and [module highlighted by blue bounding box] integrated into the model training process?
\newline
\textbf{Ground-truth Answer:} 
\newline
The question is unanswerable.

\small \textbf{Relevant Paper Context:} 
\newline
Section 3.4 \hspace{3mm} Adversarial robustness evaluation and Robustness gain:
\newline
\emph{``Adversarial perturbations are image transformations capable of fooling ANNs while remaining imperceptible for humans. To assess the adversarial robustness of our models, we employed Foolbox to create adversarial versions of the 1654 original validation images under different attack strengths.''}
\end{blueBox}

\begin{wronganswer}
\small\textbf{Qwen2-VL-72B:}\\
The EEG predictor and the image category predictor are integrated into the model training process by being connected to the ResNet50 backbone. The EEG predictor takes the output of the ResNet50 backbone and predicts the EEG signal, while the image category predictor takes the same output and predicts the image category.

\end{wronganswer}

\begin{correctanswer}
\small\textbf{Error Analysis:}\\
Qwen's answer is inaccurate because the question is explicitly marked as unanswerable, meaning the paper does not provide enough information to address it. Instead of acknowledging this, Qwen generates an unfounded explanation, incorrectly stating how the EEG predictor and image category predictor are integrated into the model training process. This response lacks support from the paper's content and contradicts the provided ground-truth answer.

\end{correctanswer}
\clearpage

%% file: appendix/error_examples/err5.tex
\subsection{Overreliance on visual elements}\label{app:err5}
\begin{blueBox}[Literature Background]
\begin{center}
    \includegraphics[width=0.8\linewidth]{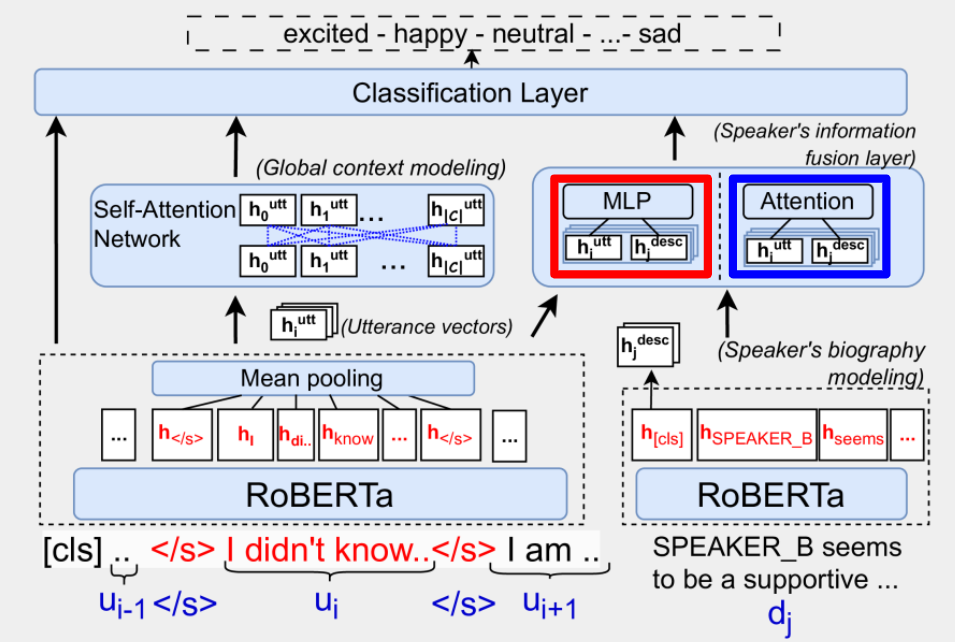}
\end{center}

\small\textbf{Paper Title:}
\newline
PS-TTL: Prototype-based Soft-labels and Test-Time Learning for Few-shot Object Detection

\textbf{Figure Caption:}
\newline
Overview of our BiosERC model architecture.

\textbf{Question:}
\newline
How are [module highlighted by red bounding box] and [module highlighted by blue bounding box] utilized to process utterance and description embeddings?

\textbf{Ground-truth Answer:} 
\newline
In BiosERC, a multi-layer perceptron (MLP) network injects personality information of speakers into their corresponding utterances, creating a unified speaker vector representation. The attention mechanism dynamically incorporates speaker information into each utterance, modeling the relationship between the utterance and all speakers in a conversation to enrich the utterance vector representation.

\small \textbf{Relevant Paper Context:} 
\newline
Section 3.4 \hspace{3mm}BERT-based BiosERC architecture:
\newline
\emph{``This architecture is designed with a straightforward target that injects the personality information of each speaker into their corresponding utterances by a multi-layer perceptron network.Through this mechanism, all the utterances from the same speaker are shared in the unified speaker vector representation, while the weights are updated in the training process. Finally, the utterance vector is fused with the speaker vector which supports emotional classification.We consider a variant of our BiosERC model, which is engineered to dynamically incorporate the speaker’s information into each utterance via the attention mechanism. The relationship between the current utterance and all individual speakers is integrated to enrich the utterance vector representation.''}

\end{blueBox}

\begin{wronganswer}
\small\textbf{Qwen2-VL-72B:}\\
The module highlighted by the red bounding box (MLP) is used to process utterance embeddings, while the module highlighted by the blue bounding box (Attention) is used to process description embeddings.

\end{wronganswer}

\begin{correctanswer}
\small\textbf{Error Analysis:}\\
Qwen2’s answer misses the core question by failing to explain how the modules process embeddings. It simply states their functions without addressing the mechanisms or their interplay, showing a lack of deeper question understanding.

\end{correctanswer}
\newpage

\begin{blueBox}[Implementation Details]
\begin{center}
    \includegraphics[width=0.8\linewidth]{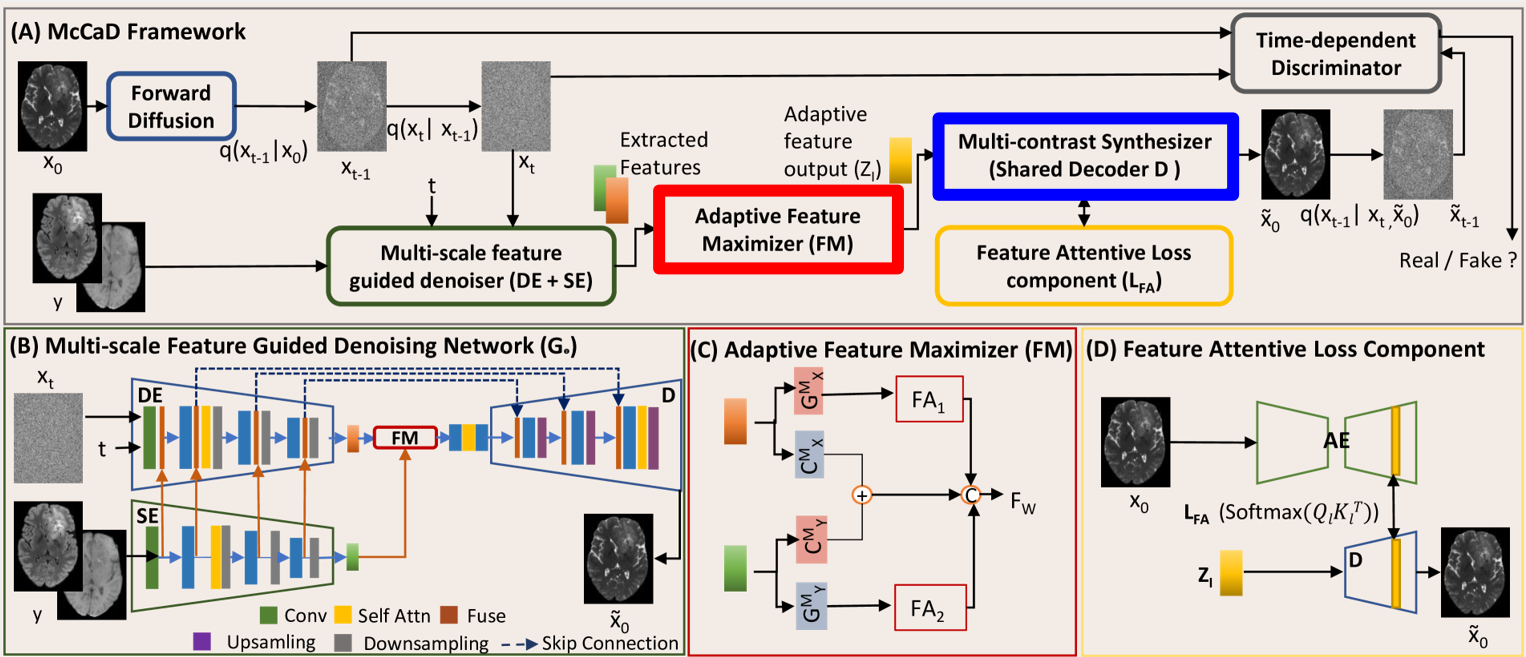}
\end{center}

\small\textbf{Paper Title:}
\newline
McCaD: Multi-Contrast MRI Conditioned, Adaptive Adversarial Diffusion Model for High-Fidelity MRI Synthesis

\textbf{Figure Caption:}
\newline
Network architecture of McCaD. A: Overall Architecture, B: Multi-scale Feature Guided Denoising Network to incorporate feature characteristics from conditional MRI contrasts at various stages to guide the reverse diffusion process, C: Adaptive Feature Maximizer, to weights more pertinent features within the latent space D: Feature Attentive Loss to improve the perceptual quality of the synthetic results.

\textbf{Question:}
\newline
What role does [module highlighted by red bounding box] play in [module highlighted by blue bounding box]?

\textbf{Ground-truth Answer:} 
\newline
The adaptive feature maximizer unifies feature distributions by utilizing encoded features from the Semantic Encoder and Diffusive Encoder, which undergo separate local and global feature extraction processes. It assigns weights based on feature relevance to facilitate the selection of mutually adaptive and effective shared representations, ultimately leading to more precise denoised target contrast.

\small \textbf{Relevant Paper Context:} 
\newline
Section 3.4 \hspace{3mm}BERT-based BiosERC architecture:
\newline
\emph{``Nevertheless, while multi-stage guidance proves beneficial in extracting valuable information from features at various levels, it is more challenging to maximize the mutual information between the conditional contrasts and the target MR contrast distributions. This is mainly due to intricate dependencies between multi-contrast imaging and finding more common and mutually adaptive feature representation.To overcome this challenge, we propose an adaptive feature maximize (
 FM) within the denoising network, unifying feature distributions as shown in Fig. 1(C).
 \newline
 The distinction between local and global feature contrasts derived from the denoising and conditional feature distributions aids in adaptively assigning weights to more pertinent features. This adaptive weighting facilitates the selection of mutually dependent and highly effective shared representations within the latent distribution. Consequently, these representations can be leveraged to achieve more precise denoised target contrast.''}

\end{blueBox}

\begin{wronganswer}
\small\textbf{Qwen2-VL-72B:}\\
The red box highlights the "Adaptive Feature Maximizer (FM)" module, and the blue box highlights the "Multi-contrast Synthesizer (Shared Decoder D)" module. The Adaptive Feature Maximizer (FM) plays a crucial role in guiding the Multi-contrast Synthesizer.

\end{wronganswer}

\begin{correctanswer}
\small\textbf{Error Analysis:}\\
Qwen's response overly relies on visual elements by merely identifying the red and blue boxes without explaining the functional relationship between FM and the synthesizer. It lacks a deeper technical interpretation of FM’s role in unifying feature distributions and guiding the synthesizer, failing to integrate the ground-truth context effectively. A stronger response should focus on the adaptive weighting mechanism and its impact on contrast precision rather than just referencing colors.

\end{correctanswer}

%% file: main.bbl
\begin{thebibliography}{38}
\expandafter\ifx\csname natexlab\endcsname\relax\def\natexlab#1{#1}\fi

\bibitem[{Abdin et~al.(2024)Abdin, Jacobs, Awan, Aneja, Awadallah, Awadalla, Bach, Bahree, Bakhtiari, Behl, Benhaim, Bilenko, Bjorck, Bubeck, Cai, Mendes, Chen, Chaudhary, Chopra, Giorno, de~Rosa, Dixon, Eldan, Iter, Garg, Goswami, Gunasekar, Haider, Hao, Hewett, Huynh, Javaheripi, Jin, Kauffmann, Karampatziakis, Kim, Khademi, Kurilenko, Lee, Lee, Li, Liang, Liu, Lin, Lin, Madan, Mitra, Modi, Nguyen, Norick, Patra, Perez-Becker, Portet, Pryzant, Qin, Radmilac, Rosset, Roy, Ruwase, Saarikivi, Saied, Salim, Santacroce, Shah, Shang, Sharma, Song, Tanaka, Wang, Ward, Wang, Witte, Wyatt, Xu, Xu, Yadav, Yang, Yang, Yu, Zhang, Zhang, Zhang, Zhang, Zhang, Zhang, Zhang, and Zhou}]{abdin2024phi3}
Marah Abdin, Sam~Ade Jacobs, Ammar~Ahmad Awan, Jyoti Aneja, Ahmed Awadallah, Hany Awadalla, Nguyen Bach, Amit Bahree, Arash Bakhtiari, Harkirat Behl, Alon Benhaim, Misha Bilenko, Johan Bjorck, Sébastien Bubeck, Martin Cai, Caio César~Teodoro Mendes, Weizhu Chen, Vishrav Chaudhary, Parul Chopra, Allie~Del Giorno, Gustavo de~Rosa, Matthew Dixon, Ronen Eldan, Dan Iter, Amit Garg, Abhishek Goswami, Suriya Gunasekar, Emman Haider, Junheng Hao, Russell~J. Hewett, Jamie Huynh, Mojan Javaheripi, Xin Jin, Piero Kauffmann, Nikos Karampatziakis, Dongwoo Kim, Mahoud Khademi, Lev Kurilenko, James~R. Lee, Yin~Tat Lee, Yuanzhi Li, Chen Liang, Weishung Liu, Eric Lin, Zeqi Lin, Piyush Madan, Arindam Mitra, Hardik Modi, Anh Nguyen, Brandon Norick, Barun Patra, Daniel Perez-Becker, Thomas Portet, Reid Pryzant, Heyang Qin, Marko Radmilac, Corby Rosset, Sambudha Roy, Olatunji Ruwase, Olli Saarikivi, Amin Saied, Adil Salim, Michael Santacroce, Shital Shah, Ning Shang, Hiteshi Sharma, Xia Song, Masahiro Tanaka, Xin Wang, Rachel
  Ward, Guanhua Wang, Philipp Witte, Michael Wyatt, Can Xu, Jiahang Xu, Sonali Yadav, Fan Yang, Ziyi Yang, Donghan Yu, Chengruidong Zhang, Cyril Zhang, Jianwen Zhang, Li~Lyna Zhang, Yi~Zhang, Yue Zhang, Yunan Zhang, and Xiren Zhou. 2024.
\newblock \href {http://arxiv.org/abs/2404.14219} {Phi-3 technical report: A highly capable language model locally on your phone}.

\bibitem[{Agrawal et~al.(2024)Agrawal, Antoniak, Hanna, Chaplot, Chudnovsky, Garg, Gervet, Ghosh, H'eliou, Jacob, Jiang, Lacroix, Lample, de~Las~Casas, Lavril, Scao, Lo, Marshall, Martin, Mensch, Muddireddy, Nemychnikova, Pellat, von Platen, Raghuraman, Rozi{\`e}re, Sablayrolles, Saulnier, Sauvestre, Shang, Soletskyi, Stewart, Stock, Studnia, Subramanian, Vaze, and Wang}]{pixtral}
Pravesh Agrawal, Szymon Antoniak, Emma~Bou Hanna, Devendra~Singh Chaplot, Jessica Chudnovsky, Saurabh Garg, Th{\'e}ophile Gervet, Soham Ghosh, Am'elie H'eliou, Paul Jacob, Albert~Q. Jiang, Timoth{\'e}e Lacroix, Guillaume Lample, Diego de~Las~Casas, Thibaut Lavril, Teven~Le Scao, Andy Lo, William Marshall, Louis Martin, Arthur Mensch, Pavankumar~Reddy Muddireddy, Valera Nemychnikova, Marie Pellat, Patrick von Platen, Nikhil Raghuraman, Baptiste Rozi{\`e}re, Alexandre Sablayrolles, Lucile Saulnier, Romain Sauvestre, Wendy Shang, Roman Soletskyi, Lawrence Stewart, Pierre Stock, Joachim Studnia, Sandeep Subramanian, Sagar Vaze, and Thomas Wang. 2024.
\newblock \href {https://api.semanticscholar.org/CorpusID:273229118} {Pixtral 12b}.
\newblock \emph{ArXiv}, abs/2410.07073.

\bibitem[{Ajith et~al.(2024)Ajith, Xia, Chevalier, Goyal, Chen, and Gao}]{ajith-etal-2024-litsearch}
Anirudh Ajith, Mengzhou Xia, Alexis Chevalier, Tanya Goyal, Danqi Chen, and Tianyu Gao. 2024.
\newblock \href {https://doi.org/10.18653/v1/2024.emnlp-main.840} {{L}it{S}earch: A retrieval benchmark for scientific literature search}.
\newblock In \emph{Proceedings of the 2024 Conference on Empirical Methods in Natural Language Processing}, pages 15068--15083, Miami, Florida, USA. Association for Computational Linguistics.

\bibitem[{Asai and Choi(2021)}]{asai-choi-2021-challenges}
Akari Asai and Eunsol Choi. 2021.
\newblock \href {https://doi.org/10.18653/v1/2021.acl-long.118} {Challenges in information-seeking {QA}: Unanswerable questions and paragraph retrieval}.
\newblock In \emph{Proceedings of the 59th Annual Meeting of the Association for Computational Linguistics and the 11th International Joint Conference on Natural Language Processing (Volume 1: Long Papers)}, pages 1492--1504, Online. Association for Computational Linguistics.

\bibitem[{Bajaj et~al.(2018)Bajaj, Campos, Craswell, Deng, Gao, Liu, Majumder, McNamara, Mitra, Nguyen, Rosenberg, Song, Stoica, Tiwary, and Wang}]{bajaj2018msmarco}
Payal Bajaj, Daniel Campos, Nick Craswell, Li~Deng, Jianfeng Gao, Xiaodong Liu, Rangan Majumder, Andrew McNamara, Bhaskar Mitra, Tri Nguyen, Mir Rosenberg, Xia Song, Alina Stoica, Saurabh Tiwary, and Tong Wang. 2018.
\newblock \href {http://arxiv.org/abs/1611.09268} {Ms marco: A human generated machine reading comprehension dataset}.

\bibitem[{Chen et~al.(2024{\natexlab{a}})Chen, Li, Dong, Zhang, Zang, Chen, Duan, Wang, Qiao, Lin, and Zhao}]{chen2024mmstar}
Lin Chen, Jinsong Li, Xiaoyi Dong, Pan Zhang, Yuhang Zang, Zehui Chen, Haodong Duan, Jiaqi Wang, Yu~Qiao, Dahua Lin, and Feng Zhao. 2024{\natexlab{a}}.
\newblock \href {http://arxiv.org/abs/2403.20330} {Are we on the right way for evaluating large vision-language models?}

\bibitem[{Chen et~al.(2024{\natexlab{b}})Chen, Wang, Cao, Liu, Gao, Cui, Zhu, Ye, Tian, Liu et~al.}]{chen2024expanding}
Zhe Chen, Weiyun Wang, Yue Cao, Yangzhou Liu, Zhangwei Gao, Erfei Cui, Jinguo Zhu, Shenglong Ye, Hao Tian, Zhaoyang Liu, et~al. 2024{\natexlab{b}}.
\newblock Expanding performance boundaries of open-source multimodal models with model, data, and test-time scaling.
\newblock \emph{arXiv preprint arXiv:2412.05271}.

\bibitem[{Chen et~al.(2024{\natexlab{c}})Chen, Wang, Tian, Ye, Gao, Cui, Tong, Hu, Luo, Ma et~al.}]{chen2024far}
Zhe Chen, Weiyun Wang, Hao Tian, Shenglong Ye, Zhangwei Gao, Erfei Cui, Wenwen Tong, Kongzhi Hu, Jiapeng Luo, Zheng Ma, et~al. 2024{\natexlab{c}}.
\newblock How far are we to gpt-4v? closing the gap to commercial multimodal models with open-source suites.
\newblock \emph{arXiv preprint arXiv:2404.16821}.

\bibitem[{Chen et~al.(2023)Chen, Wu, Wang, Su, Chen, Xing, Zhong, Zhang, Zhu, Lu, Li, Luo, Lu, Qiao, and Dai}]{chen2023internvl}
Zhe Chen, Jiannan Wu, Wenhai Wang, Weijie Su, Guo Chen, Sen Xing, Muyan Zhong, Qinglong Zhang, Xizhou Zhu, Lewei Lu, Bin Li, Ping Luo, Tong Lu, Yu~Qiao, and Jifeng Dai. 2023.
\newblock Internvl: Scaling up vision foundation models and aligning for generic visual-linguistic tasks.
\newblock \emph{arXiv preprint arXiv:2312.14238}.

\bibitem[{Dasigi et~al.(2021)Dasigi, Lo, Beltagy, Cohan, Smith, and Gardner}]{dasigi-etal-2021-dataset}
Pradeep Dasigi, Kyle Lo, Iz~Beltagy, Arman Cohan, Noah~A. Smith, and Matt Gardner. 2021.
\newblock \href {https://doi.org/10.18653/v1/2021.naacl-main.365} {A dataset of information-seeking questions and answers anchored in research papers}.
\newblock In \emph{Proceedings of the 2021 Conference of the North American Chapter of the Association for Computational Linguistics: Human Language Technologies}, pages 4599--4610, Online. Association for Computational Linguistics.

\bibitem[{Deng et~al.(2024)Deng, Zhao, Heng, Li, Cao, Tang, and Cohan}]{deng-etal-2024-unveiling}
Chunyuan Deng, Yilun Zhao, Yuzhao Heng, Yitong Li, Jiannan Cao, Xiangru Tang, and Arman Cohan. 2024.
\newblock \href {https://doi.org/10.18653/v1/2024.findings-acl.951} {Unveiling the spectrum of data contamination in language model: A survey from detection to remediation}.
\newblock In \emph{Findings of the Association for Computational Linguistics: ACL 2024}, pages 16078--16092, Bangkok, Thailand. Association for Computational Linguistics.

\bibitem[{Gemini(2024)}]{geminiteam2024gemini}
Gemini. 2024.
\newblock \href {http://arxiv.org/abs/2403.05530} {Gemini 1.5: Unlocking multimodal understanding across millions of tokens of context}.

\bibitem[{Jiang et~al.(2023)Jiang, Sablayrolles, Mensch, Bamford, Chaplot, de~Las~Casas, Bressand, Lengyel, Lample, Saulnier, Lavaud, Lachaux, Stock, Scao, Lavril, Wang, Lacroix, and Sayed}]{Jiang2023Mistral7}
Albert~Qiaochu Jiang, Alexandre Sablayrolles, Arthur Mensch, Chris Bamford, Devendra~Singh Chaplot, Diego de~Las~Casas, Florian Bressand, Gianna Lengyel, Guillaume Lample, Lucile Saulnier, L'elio~Renard Lavaud, Marie-Anne Lachaux, Pierre Stock, Teven~Le Scao, Thibaut Lavril, Thomas Wang, Timoth{\'e}e Lacroix, and William~El Sayed. 2023.
\newblock \href {https://api.semanticscholar.org/CorpusID:263830494} {Mistral 7b}.
\newblock \emph{ArXiv}, abs/2310.06825.

\bibitem[{Lee et~al.(2023{\natexlab{a}})Lee, Lee, Park, Hwang, Kim, Lee, and Lee}]{qasa}
Yoonjoo Lee, Kyungjae Lee, Sunghyun Park, Dasol Hwang, Jaehyeon Kim, Hong-In Lee, and Moontae Lee. 2023{\natexlab{a}}.
\newblock \href {https://proceedings.mlr.press/v202/lee23n.html} {{QASA}: Advanced question answering on scientific articles}.
\newblock In \emph{Proceedings of the 40th International Conference on Machine Learning}, volume 202 of \emph{Proceedings of Machine Learning Research}, pages 19036--19052. PMLR.

\bibitem[{Lee et~al.(2023{\natexlab{b}})Lee, Lee, Park, Hwang, Kim, Lee, and Lee}]{pmlr-v202-lee23n}
Yoonjoo Lee, Kyungjae Lee, Sunghyun Park, Dasol Hwang, Jaehyeon Kim, Hong-In Lee, and Moontae Lee. 2023{\natexlab{b}}.
\newblock \href {https://proceedings.mlr.press/v202/lee23n.html} {{QASA}: Advanced question answering on scientific articles}.
\newblock In \emph{Proceedings of the 40th International Conference on Machine Learning}, volume 202 of \emph{Proceedings of Machine Learning Research}, pages 19036--19052. PMLR.

\bibitem[{Li et~al.(2024{\natexlab{a}})Li, Shangguan, Zhao, Li, Liu, and Cohan}]{li-etal-2024-m3sciqa}
Chuhan Li, Ziyao Shangguan, Yilun Zhao, Deyuan Li, Yixin Liu, and Arman Cohan. 2024{\natexlab{a}}.
\newblock \href {https://doi.org/10.18653/v1/2024.findings-emnlp.904} {{M}3{S}ci{QA}: A multi-modal multi-document scientific {QA} benchmark for evaluating foundation models}.
\newblock In \emph{Findings of the Association for Computational Linguistics: EMNLP 2024}, pages 15419--15446, Miami, Florida, USA. Association for Computational Linguistics.

\bibitem[{Li et~al.(2024{\natexlab{b}})Li, Wang, Xu, Wang, Feng, Kong, and Liu}]{li-etal-2024-multimodal-arxiv}
Lei Li, Yuqi Wang, Runxin Xu, Peiyi Wang, Xiachong Feng, Lingpeng Kong, and Qi~Liu. 2024{\natexlab{b}}.
\newblock \href {https://doi.org/10.18653/v1/2024.acl-long.775} {Multimodal {A}r{X}iv: A dataset for improving scientific comprehension of large vision-language models}.
\newblock In \emph{Proceedings of the 62nd Annual Meeting of the Association for Computational Linguistics (Volume 1: Long Papers)}, pages 14369--14387, Bangkok, Thailand. Association for Computational Linguistics.

\bibitem[{Li et~al.(2024{\natexlab{c}})Li, Wang, Xu, Wang, Feng, Kong, and Liu}]{li2024multimodal}
Lei Li, Yuqi Wang, Runxin Xu, Peiyi Wang, Xiachong Feng, Lingpeng Kong, and Qi~Liu. 2024{\natexlab{c}}.
\newblock \href {http://arxiv.org/abs/2403.00231} {Multimodal arxiv: A dataset for improving scientific comprehension of large vision-language models}.

\bibitem[{Li et~al.(2024{\natexlab{d}})Li, Yang, Choi, Zhu, Hsieh, Kim, Lim, Ji, Lee, Yan, Petzold, Wilson, Lim, and Wang}]{li2024mmsci}
Zekun Li, Xianjun Yang, Kyuri Choi, Wanrong Zhu, Ryan Hsieh, HyeonJung Kim, Jin~Hyuk Lim, Sungyoung Ji, Byungju Lee, Xifeng Yan, Linda~Ruth Petzold, Stephen~D. Wilson, Woosang Lim, and William~Yang Wang. 2024{\natexlab{d}}.
\newblock \href {http://arxiv.org/abs/2407.04903} {Mmsci: A dataset for graduate-level multi-discipline multimodal scientific understanding}.

\bibitem[{Lu et~al.(2024)Lu, Bansal, Xia, Liu, Li, Hajishirzi, Cheng, Chang, Galley, and Gao}]{lu2024mathvista}
Pan Lu, Hritik Bansal, Tony Xia, Jiacheng Liu, Chunyuan Li, Hannaneh Hajishirzi, Hao Cheng, Kai-Wei Chang, Michel Galley, and Jianfeng Gao. 2024.
\newblock Mathvista: Evaluating mathematical reasoning of foundation models in visual contexts.
\newblock In \emph{International Conference on Learning Representations (ICLR)}.

\bibitem[{Microsoft et~al.(2025)Microsoft, :, Abouelenin, Ashfaq, Atkinson, Awadalla, Bach, Bao, Benhaim, Cai, Chaudhary, Chen, Chen, Chen, Chen, Chen, Chen, ling Chen, Dai, Dai, Fan, Gao, Gao, Garg, Goswami, Hao, Hendy, Hu, Jin, Khademi, Kim, Kim, Lee, Li, Li, Liang, Lin, Lin, Liu, Liu, Lopez, Luo, Madan, Mazalov, Mitra, Mousavi, Nguyen, Pan, Perez-Becker, Platin, Portet, Qiu, Ren, Ren, Roy, Shang, Shen, Singhal, Som, Song, Sych, Vaddamanu, Wang, Wang, Wang, Wu, Xu, Xu, Yang, Yang, Yu, Zabir, Zhang, Zhang, Zhang, and Zhou}]{microsoft2025phi4}
Microsoft, :, Abdelrahman Abouelenin, Atabak Ashfaq, Adam Atkinson, Hany Awadalla, Nguyen Bach, Jianmin Bao, Alon Benhaim, Martin Cai, Vishrav Chaudhary, Congcong Chen, Dong Chen, Dongdong Chen, Junkun Chen, Weizhu Chen, Yen-Chun Chen, Yi~ling Chen, Qi~Dai, Xiyang Dai, Ruchao Fan, Mei Gao, Min Gao, Amit Garg, Abhishek Goswami, Junheng Hao, Amr Hendy, Yuxuan Hu, Xin Jin, Mahmoud Khademi, Dongwoo Kim, Young~Jin Kim, Gina Lee, Jinyu Li, Yunsheng Li, Chen Liang, Xihui Lin, Zeqi Lin, Mengchen Liu, Yang Liu, Gilsinia Lopez, Chong Luo, Piyush Madan, Vadim Mazalov, Arindam Mitra, Ali Mousavi, Anh Nguyen, Jing Pan, Daniel Perez-Becker, Jacob Platin, Thomas Portet, Kai Qiu, Bo~Ren, Liliang Ren, Sambuddha Roy, Ning Shang, Yelong Shen, Saksham Singhal, Subhojit Som, Xia Song, Tetyana Sych, Praneetha Vaddamanu, Shuohang Wang, Yiming Wang, Zhenghao Wang, Haibin Wu, Haoran Xu, Weijian Xu, Yifan Yang, Ziyi Yang, Donghan Yu, Ishmam Zabir, Jianwen Zhang, Li~Lyna Zhang, Yunan Zhang, and Xiren Zhou. 2025.
\newblock \href {http://arxiv.org/abs/2503.01743} {Phi-4-mini technical report: Compact yet powerful multimodal language models via mixture-of-loras}.

\bibitem[{{Mistral AI}(2025)}]{mistral2025small31}
{Mistral AI}. 2025.
\newblock Mistral-small-3.1-24b-instruct-2503.
\newblock \url{https://huggingface.co/mistralai/Mistral-Small-3.1-24B-Instruct-2503}.
\newblock Apache 2.0 License.

\bibitem[{OpenAI(2024)}]{openai2024gpt4o}
OpenAI. 2024.
\newblock \href {https://openai.com/index/hello-gpt-4o/} {Hello gpt-4o}.

\bibitem[{OpenAI(2025{\natexlab{a}})}]{openai2025o4mini}
OpenAI. 2025{\natexlab{a}}.
\newblock \href {https://openai.com/index/o3-o4-mini-system-card-addendum-operator-o3/} {Addendum to openai o3 and o4-mini system card: Openai o3 operator}.

\bibitem[{OpenAI(2025{\natexlab{b}})}]{openai2024gpt4-1}
OpenAI. 2025{\natexlab{b}}.
\newblock \href {https://openai.com/index/gpt-4-1/} {Introducing gpt-4.1 in the api}.

\bibitem[{Rajpurkar et~al.(2018)Rajpurkar, Jia, and Liang}]{rajpurkar-etal-2018-know}
Pranav Rajpurkar, Robin Jia, and Percy Liang. 2018.
\newblock \href {https://doi.org/10.18653/v1/P18-2124} {Know what you don{'}t know: Unanswerable questions for {SQ}u{AD}}.
\newblock In \emph{Proceedings of the 56th Annual Meeting of the Association for Computational Linguistics (Volume 2: Short Papers)}, pages 784--789, Melbourne, Australia. Association for Computational Linguistics.

\bibitem[{Rajpurkar et~al.(2016)Rajpurkar, Zhang, Lopyrev, and Liang}]{rajpurkar2016squad}
Pranav Rajpurkar, Jian Zhang, Konstantin Lopyrev, and Percy Liang. 2016.
\newblock \href {http://arxiv.org/abs/1606.05250} {Squad: 100,000+ questions for machine comprehension of text}.

\bibitem[{Sulem et~al.(2022)Sulem, Hay, and Roth}]{sulem-etal-2022-yes}
Elior Sulem, Jamaal Hay, and Dan Roth. 2022.
\newblock \href {https://doi.org/10.18653/v1/2022.naacl-main.79} {Yes, no or {IDK}: The challenge of unanswerable yes/no questions}.
\newblock In \emph{Proceedings of the 2022 Conference of the North American Chapter of the Association for Computational Linguistics: Human Language Technologies}, pages 1075--1085, Seattle, United States. Association for Computational Linguistics.

\bibitem[{Touvron et~al.(2023)Touvron, Martin, Stone, Albert, Almahairi, Babaei, Bashlykov, Batra, Bhargava, Bhosale, Bikel, Blecher, Ferrer, Chen, Cucurull, Esiobu, Fernandes, Fu, Fu, Fuller, Gao, Goswami, Goyal, Hartshorn, Hosseini, Hou, Inan, Kardas, Kerkez, Khabsa, Kloumann, Korenev, Koura, Lachaux, Lavril, Lee, Liskovich, Lu, Mao, Martinet, Mihaylov, Mishra, Molybog, Nie, Poulton, Reizenstein, Rungta, Saladi, Schelten, Silva, Smith, Subramanian, Tan, Tang, Taylor, Williams, Kuan, Xu, Yan, Zarov, Zhang, Fan, Kambadur, Narang, Rodriguez, Stojnic, Edunov, and Scialom}]{Touvron2023Llama2O}
Hugo Touvron, Louis Martin, Kevin~R. Stone, Peter Albert, Amjad Almahairi, Yasmine Babaei, Nikolay Bashlykov, Soumya Batra, Prajjwal Bhargava, Shruti Bhosale, Daniel~M. Bikel, Lukas Blecher, Cristian~Cant{\'o}n Ferrer, Moya Chen, Guillem Cucurull, David Esiobu, Jude Fernandes, Jeremy Fu, Wenyin Fu, Brian Fuller, Cynthia Gao, Vedanuj Goswami, Naman Goyal, Anthony~S. Hartshorn, Saghar Hosseini, Rui Hou, Hakan Inan, Marcin Kardas, Viktor Kerkez, Madian Khabsa, Isabel~M. Kloumann, A.~V. Korenev, Punit~Singh Koura, Marie-Anne Lachaux, Thibaut Lavril, Jenya Lee, Diana Liskovich, Yinghai Lu, Yuning Mao, Xavier Martinet, Todor Mihaylov, Pushkar Mishra, Igor Molybog, Yixin Nie, Andrew Poulton, Jeremy Reizenstein, Rashi Rungta, Kalyan Saladi, Alan Schelten, Ruan Silva, Eric~Michael Smith, R.~Subramanian, Xia Tan, Binh Tang, Ross Taylor, Adina Williams, Jian~Xiang Kuan, Puxin Xu, Zhengxu Yan, Iliyan Zarov, Yuchen Zhang, Angela Fan, Melanie Kambadur, Sharan Narang, Aurelien Rodriguez, Robert Stojnic, Sergey Edunov, and
  Thomas Scialom. 2023.
\newblock \href {https://api.semanticscholar.org/CorpusID:259950998} {Llama 2: Open foundation and fine-tuned chat models}.
\newblock \emph{ArXiv}, abs/2307.09288.

\bibitem[{Wang et~al.(2025)Wang, Shen, Kuang, Cohan, and Zhao}]{wang2025sciver}
Chengye Wang, Yifei Shen, Zexi Kuang, Arman Cohan, and Yilun Zhao. 2025.
\newblock \href {http://arxiv.org/abs/2506.15569} {Sciver: Evaluating foundation models for multimodal scientific claim verification}.

\bibitem[{Wang et~al.(2024{\natexlab{a}})Wang, Bai, Tan, Wang, Fan, Bai, Chen, Liu, Wang, Ge, Fan, Dang, Du, Ren, Men, Liu, Zhou, Zhou, and Lin}]{wang2024qwen2vl}
Peng Wang, Shuai Bai, Sinan Tan, Shijie Wang, Zhihao Fan, Jinze Bai, Keqin Chen, Xuejing Liu, Jialin Wang, Wenbin Ge, Yang Fan, Kai Dang, Mengfei Du, Xuancheng Ren, Rui Men, Dayiheng Liu, Chang Zhou, Jingren Zhou, and Junyang Lin. 2024{\natexlab{a}}.
\newblock \href {http://arxiv.org/abs/2409.12191} {Qwen2-vl: Enhancing vision-language model's perception of the world at any resolution}.

\bibitem[{Wang et~al.(2024{\natexlab{b}})Wang, Xia, He, Chen, Liu, Zhu, Liang, Wu, Liu, Malladi, Chevalier, Arora, and Chen}]{wang2024charxiv}
Zirui Wang, Mengzhou Xia, Luxi He, Howard Chen, Yitao Liu, Richard Zhu, Kaiqu Liang, Xindi Wu, Haotian Liu, Sadhika Malladi, Alexis Chevalier, Sanjeev Arora, and Danqi Chen. 2024{\natexlab{b}}.
\newblock \href {http://arxiv.org/abs/2406.18521} {Charxiv: Charting gaps in realistic chart understanding in multimodal llms}.

\bibitem[{Xu et~al.(2025)Xu, Zhao, Patwardhan, Vig, and Cohan}]{xu2025llmsidentifycriticallimitations}
Zhijian Xu, Yilun Zhao, Manasi Patwardhan, Lovekesh Vig, and Arman Cohan. 2025.
\newblock \href {http://arxiv.org/abs/2507.02694} {Can llms identify critical limitations within scientific research? a systematic evaluation on ai research papers}.

\bibitem[{Yang et~al.(2018)Yang, Qi, Zhang, Bengio, Cohen, Salakhutdinov, and Manning}]{yang2018hotpotqa}
Zhilin Yang, Peng Qi, Saizheng Zhang, Yoshua Bengio, William~W. Cohen, Ruslan Salakhutdinov, and Christopher~D. Manning. 2018.
\newblock \href {http://arxiv.org/abs/1809.09600} {Hotpotqa: A dataset for diverse, explainable multi-hop question answering}.

\bibitem[{Yue et~al.(2024)Yue, Zheng, Ni, Wang, Zhang, Tong, Sun, Yu, Zhang, Sun, Su, Chen, and Neubig}]{yue2024mmmupro}
Xiang Yue, Tianyu Zheng, Yuansheng Ni, Yubo Wang, Kai Zhang, Shengbang Tong, Yuxuan Sun, Botao Yu, Ge~Zhang, Huan Sun, Yu~Su, Wenhu Chen, and Graham Neubig. 2024.
\newblock \href {http://arxiv.org/abs/2409.02813} {Mmmu-pro: A more robust multi-discipline multimodal understanding benchmark}.

\bibitem[{Zhang et~al.(2024)Zhang, Jiang, Zhang, Lin, Guo, Qiu, Zhou, Lu, Chang, Gao, and Li}]{zhang2024mathverse}
Renrui Zhang, Dongzhi Jiang, Yichi Zhang, Haokun Lin, Ziyu Guo, Pengshuo Qiu, Aojun Zhou, Pan Lu, Kai-Wei Chang, Peng Gao, and Hongsheng Li. 2024.
\newblock \href {http://arxiv.org/abs/2403.14624} {Mathverse: Does your multi-modal llm truly see the diagrams in visual math problems?}

\bibitem[{Zhao et~al.(2025{\natexlab{a}})Zhao, Zhang, Xie, Hu, Gan, Long, Hu, Chen, Li, Xu, Wang, Shangguan, Liang, Liu, Zhao, and Cohan}]{Zhao_2025_CVPR}
Yilun Zhao, Haowei Zhang, Lujing Xie, Tongyan Hu, Guo Gan, Yitao Long, Zhiyuan Hu, Weiyuan Chen, Chuhan Li, Zhijian Xu, Chengye Wang, Ziyao Shangguan, Zhenwen Liang, Yixin Liu, Chen Zhao, and Arman Cohan. 2025{\natexlab{a}}.
\newblock Mmvu: Measuring expert-level multi-discipline video understanding.
\newblock In \emph{Proceedings of the Computer Vision and Pattern Recognition Conference (CVPR)}, pages 8475--8489.

\bibitem[{Zhao et~al.(2025{\natexlab{b}})Zhao, Zhang, Hu, Wu, Bras, Anderson, Bragg, Chang, Dodge, Latzke, Liu, McGrady, Tang, Wang, Zhao, Hajishirzi, Downey, and Cohan}]{zhao2025sciarena}
Yilun Zhao, Kaiyan Zhang, Tiansheng Hu, Sihong Wu, Ronan~Le Bras, Taira Anderson, Jonathan Bragg, Joseph~Chee Chang, Jesse Dodge, Matt Latzke, Yixin Liu, Charles McGrady, Xiangru Tang, Zihang Wang, Chen Zhao, Hannaneh Hajishirzi, Doug Downey, and Arman Cohan. 2025{\natexlab{b}}.
\newblock \href {http://arxiv.org/abs/2507.01001} {Sciarena: An open evaluation platform for foundation models in scientific literature tasks}.

\end{thebibliography}
